\pdfoutput=1
\documentclass[11pt]{article}

\usepackage[preprint]{acl}

\usepackage{times}
\usepackage{latexsym}
\usepackage[T1]{fontenc}
\usepackage[utf8]{inputenc}
\usepackage{microtype}
\usepackage{inconsolata}
\usepackage{bm}
\usepackage{amsmath}
\usepackage{graphicx}
\usepackage{float}
\usepackage{multirow}
\usepackage{epstopdf}
\usepackage{algorithmic}
\usepackage{algorithm}
\usepackage{xcolor}
\usepackage{tcolorbox}
\usepackage{booktabs}
\usepackage{bbm}
\usepackage{url}
\usepackage{siunitx}
\usepackage{colortbl}
\usepackage{amssymb}

\usepackage[utf8]{inputenc}
\usepackage{CJKutf8}

\usepackage{subcaption}

\title{Zero-shot Cross-lingual NER via Mitigating Language Difference: An Entity-aligned Translation Perspective}

\author{Zhihao Zhang$^1$, Sophia Yat Mei Lee$^2$, Dong Zhang$^1$\footnotemark[1], {\bf Shoushan Li$^1$} \and {\bf Guodong Zhou$^1$}\\
$^1$School of Computer Science \& Technology, NLP Lab, Soochow University, China \\
$^2$Department of Chinese and Bilingual Studies, The Hong Kong Polytechnic University \\
\texttt{dzhang@suda.edu.cn}
}

\begin{document}
\begin{CJK}{UTF8}{gkai}

\maketitle
\renewcommand{\thefootnote}{\fnsymbol{footnote}}
\footnotetext[1]{Corresponding Author}
\begin{abstract}
Cross-lingual Named Entity Recognition (CL-NER) aims to transfer knowledge from high-resource languages to low-resource languages. However, existing zero-shot CL-NER (ZCL-NER) approaches primarily focus on Latin script language (LSL), where shared linguistic features facilitate effective knowledge transfer. In contrast, for non-Latin script language (NSL), such as Chinese and Japanese, performance often degrades due to deep structural differences. To address these challenges, we propose an entity-aligned translation (\textsc{Eat}) approach~\footnotemark[2]. Leveraging large language models (LLMs), \textsc{Eat} employs a dual-translation strategy to align entities between NSL and English. In addition, we fine-tune LLMs using multilingual Wikipedia data to enhance the entity alignment from source to target languages.
% Our approach enhances ZCL-NER performance on NSLs . 
Extensive experiments demonstrate that \textsc{Eat} outperforms prior methods on NSL by bridging language gaps through entity-aware translation.

\footnotetext[2]{Our code and dataset are available at: \url{https://github.com/ZelateCalcite/EAT\_NER} }
\end{abstract}

\section{Introduction}
% [E.G. 中文语法规则虽然复杂，但是常用语序以SVO较多，日语则是SOV。recent 日本社媒流行伪中国语对话，即去除日语假名，仅以汉字进行交流，由于去除假名导致日语的词失去了语法结构（黏着语特性），而使得伪中国语的语法结构由日语的SOV变为中文更常用的SVO。同时伪中国语可以交流的一个主要原因是汉字本身包含的大量信息，即使抛弃语法也能在一定程度上传达出说话者想表达的含义]

% Grammar and lexicon共同表达语义，其中部分grammar可以是通过一些介词来体现时态等详细语义。省略这些介词会遗失时态信息，但是句子承载的大部分语义还是通过lexicon保留了下来。在拥有重叠词表的语言中，这可以帮助跨语言的语义理解。

% Named Entity Recognition (NER) is a fundamental natural language processing (NLP) task that involves identifying and classifying named entities (e.g. Person, Location, etc.) in a text sequence. Previous deep learning-based NER approaches have achieved state-of-the-art (SOTA) results. However, they depend on large amounts of manually annotated data~\cite{li2021acl, mhaske2023acl}, and they are neutralized in low-resource scenarios~\cite{xie2023emnlp, xie2024acl}.
% On this basis, cross-lingual NER (CLNER) aims to transfer knowledge from high-resource language to low-resource language~\cite{ProKD, DenKD}.

Cross-lingual Named Entity Recognition (CL-NER) aims to transfer knowledge from high-resource source languages to low-resource target languages, so as to enhance the NER performance ~\cite{li2021acl, mhaske2023acl,xie2023emnlp, xie2024acl}. Recently, while zero-shot CL-NER (ZCL-NER) approaches demonstrate strong performance on several low-resource languages~\cite{DualNER, ProKD, DenKD}, we observe an interesting phenomenon in prior approaches and different language scripts.
% between the latin script and non-latin script target languages.

% \begin{figure}[t]
%   \centering
%   \includegraphics[width=2.4in]{intro.pdf}
%   \caption{Two examples: German, also as LSL, tends to be translated more accurately into English due to shared lexicon, making it more suitable for NER. While, Chinese, as NSL, faces inherent challenges in translation to English because of significant typological divergences.}
%   \label{fig:lexicon}
% \end{figure}

% \begin{figure}[t]
%   \centering
%   \includegraphics[width=\linewidth]{intro.pdf}
%   \caption{Two examples: German, also as LSL, tends to be translated more accurately into English due to shared lexicon, making it more suitable for NER. While, Chinese, as NSL, faces inherent challenges in translation to English because of significant typological divergences.}
%   \label{fig:lexicon}
% \end{figure}

\begin{figure}[t]
  \centering
  \includegraphics[width=\linewidth]{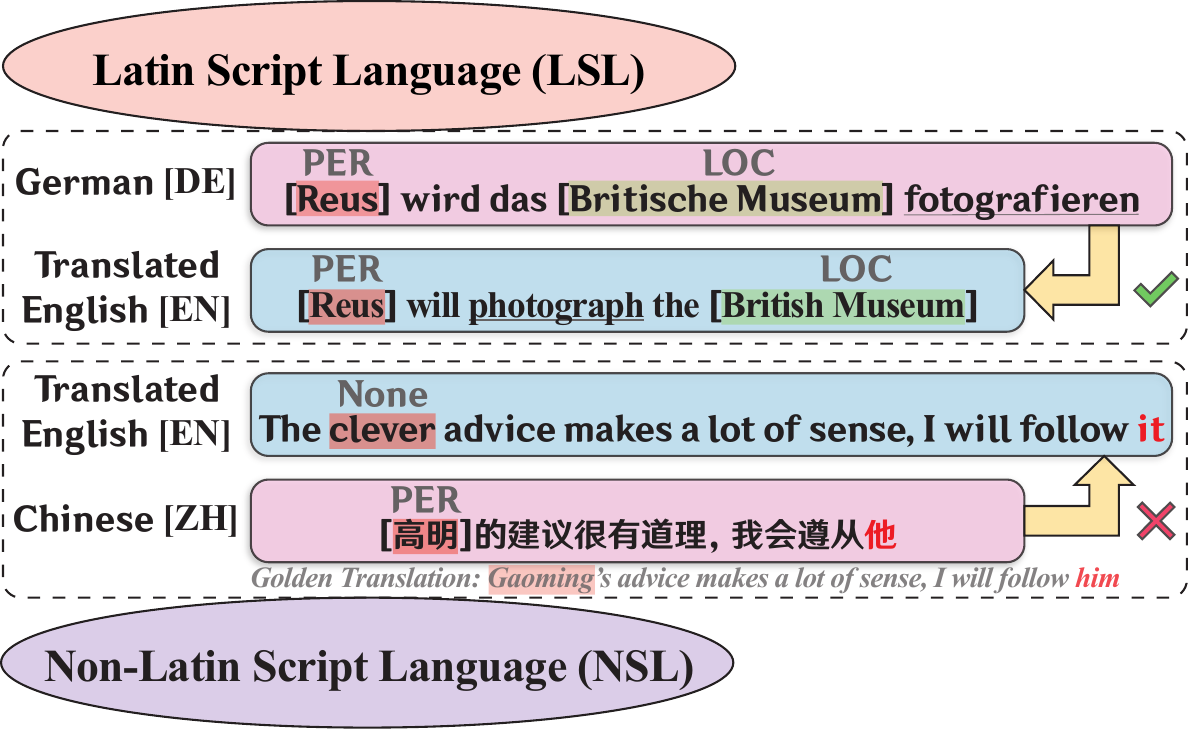}
  \caption{Two examples: German, as LSL, tends to be translated more accurately into English due to their shared lexicon, making it more suitable for NER. In contrast, Chinese, as NSL, faces inherent challenges in translation to English because of significant typological divergences. The translations are obtained by GPT-4.}
  \label{fig:lexicon}
\end{figure}

% .   by focusing on English as source language due to the popularity and high resource . 
\begin{table*}[ht]
\centering
\small
  \resizebox{\textwidth}{!}{
  \begin{tabular}{l|cccc|cc|c}
     \toprule
     {} & \textbf{AR}& \textbf{HI} & \textbf{HY} & \textbf{JA} & \ \ \textbf{ES}\ \  & \textbf{FR} & \textbf{EN} \\
     \midrule
     \multirow{2}{*}{Language Family} & {Afroasiatic} & {Indo-European} & {Indo-European} & {Japonic} & \multicolumn{2}{c|}{Indo-European} & \cellcolor{red!10}{Indo-European} \\
     
      & {$\cdot$ Semitic} & {$\cdot$ Indo-Iranian} & {$\cdot$ Armenian} & {$\cdot$ Japanese} & \multicolumn{2}{c|}{$\cdot$ Romance} & \cellcolor{red!10}{$\cdot$ Germanic} \\

     \rowcolor{gray!10}
     {Linguistic Type} & {Fusional} & {Fusional} & {Fusional} & {Agglutinative} & \multicolumn{2}{c|}{Fusional} & {Analytic}\\
     
     {Scripts} & {Arabic Abjad} & {Devanagari} & {Armenian script} & {Kanji \& Kana} & \multicolumn{2}{c|}{Latin Scripts$^{\romannumeral1}$} & \cellcolor{red!10}{Latin Scripts$^{\romannumeral1}$}\\
     
     \rowcolor{gray!10}
     {Word Order} & VSO$^{\romannumeral2}$ & SOV & SOV & SOV & \multicolumn{2}{c|}{SVO} & SVO \\
     \midrule
     {} & \textbf{KA} & \textbf{KO} & \textbf{RU} & \textbf{ZH} & \textbf{DE} & \textbf{NL} & \textbf{EN} \\
     \midrule
     \multirow{2}{*}{Language Family} & {Kartvelian} & {Koreanic} & {Indo-European} & {Sino-Tibetan} & \multicolumn{2}{c|}{Indo-European} & \cellcolor{red!10}{Indo-European} \\
      & {$\cdot$ Karto-Zan} & {$\cdot$ Korean} & {$\cdot$ Slavic} &
     {$\cdot$ Sinitic} & \multicolumn{2}{c|}{$\cdot$ Germanic} & \cellcolor{red!10}{$\cdot$ Germanic} \\
     
     \rowcolor{gray!10}
     {Linguistic Type} & {Agglutinative} & {Agglutinative} & {Fusional} & {Isolating} & \multicolumn{2}{c|}{Fusional} & {Analytic} \\

     {Scripts} & {Georgian Scripts} & {Hangul / Chosŏn'gŭl} & {Cyrillic} & {Chinese Characters} & \multicolumn{2}{c|}{Latin Scripts$^{\romannumeral1}$} & \cellcolor{red!10}{Latin Scripts$^{\romannumeral1}$} \\
     
     \rowcolor{gray!10}
     {Word Order} & SVO & SOV & SVO & SVO & \multicolumn{2}{c|}{SVO$^{\romannumeral3}$} & SVO \\
     \bottomrule
  \end{tabular}
  }
  \caption{Comparison of linguistic typology between different languages using different scripts (Language Code follows ISO 639-1:2002~\protect\footnotemark[1]). SVO (Subject-Verb-Object, similar with SOV / VSO) refers to the order in which the elements of a sentence typically appear in languages that follow this structure. $^{\romannumeral1}$: These languages are both based on Latin Scripts with unique pronunciation differences or additional characters. $^{\romannumeral2}$: The word order of Modern Written Arabic is VSO while Modern Spoken Arabic is SVO, here we only discuss the writing systems. $^{\romannumeral3}$: The usual word order of these two languages is SVO, but in subordinate clauses the word order shifts to SOV.}
  \label{tab:lang-prop}
\end{table*}

Previous ZCL-NER approaches typically apply a teacher-student (T-S) learning framework, transferring English knowledge to the target language \cite{MSD,DualNER}.
This recasts the \textbf{source language into the same space as the target language} to achieve ZCL-NER.
Regarding the existing studies~\cite{ProKD,DenKD}, they achieve competitive performance on the English-like target languages that have shared vocabulary origins, as well as similar grammatical and syntactic structures such as German and French~\cite{FINIASZ2024105707}. For example, as illustrated in Figure~\ref{fig:lexicon}, the German words ``Britische'' and ``fotografieren'' are derived from the English words ``British'' and ``photograph'', respectively. We refer to such languages, which are closely related to English, as Latin script language (\textbf{LSL}). However, as shown in our pilot experiments and previous reports \cite{ProKD,DenKD}, T-S approaches work not well on non-Latin script language (\textbf{NSL}), such as Chinese and Japanese. This is mainly due to significant linguistic discrepancies between LSL and NSL, i.e., differences in scripts, grammar, and syntax \cite{Singh2022}, which are summarized in Table~\ref{tab:lang-prop} for various languages. For example, German (DE), as LSL, shares the same Subject-Verb-Object (SVO) and fusional scripts characteristics with English (EN), while Japanese (JA), as NSL, is an agglutinative language with Subject-Object-Verb (SOV) order. Consequently, translating Japanese into English poses greater challenges compared to German due to these typological divergences.

To this end, we believe that it is essential to mitigate the language gap between NSL and English. In this way, we can utilize the abundant English resources to improve ZCL-NER performance on targeted NSL, which constitutes the core focus of this paper. As we know, translation appears to be the most intuitive approach to bridging the linguistic disparity~\cite{ijcai/LiHCCGZ24}. However, direct translation between English and NSL typically results in the key entity omissions due to various word misalignment issues~\cite{crop}.
% For example, in Figure~\ref{fig:lexicon}, the Person entity ``李美丽'' in target language Chinese is translated as a name ``Li'' and an adverb ``beautifully''. This makes our task impossible to proceed.
For instance, in Figure~\ref{fig:lexicon}, the Person entity ``高明'' in target language (Chinese) is incorrectly translated as the adjective ``clever'' (by GPT-4o and Deepseek in May 2025 with evidences in Appendix). Such translation inconsistencies hinder the progress of our task.
% , in which entities are actually fundamental in our NER task. , which absolutely can not be recognized in target language.

\footnotetext[1]{\url{https://en.m.wikipedia.org/wiki/ISO_639-1}}

To address the above issues, we propose an Entity-Aligned Translation (\textsc{Eat}) approach at dual levels with large language models (LLMs) for ZCL-NER, with a focus on NSL as the target languages.
Different from T-S approaches, we recast \textbf{the target NSL into the same space with English} to achieve ZCL-NER. 
% In other words, our \textsc{Eat} starts from the perspective of improving the ability to translate NSL into entity-aligned English to mitigate the language difference.
Specifically, we first leverage the powerful reasoning and interpretation abilities of LLMs to perform target-to-English forward translation using multi-round chain-of-thought (MrCoT). Then, we extract potential entities from the translated English text using pre-trained English NER extractor. Finally, to restore these entities in the target language, we design a backward translation process with MrCoT, ensuring that this stage's translated entities correspond to the correct fragments in the original sentence.
% At this stage, it is necessary to check whether the translated target language entity belongs to a fragment in the original sentence of the target language. 
The process allows us to achieve ZCL-NER without relying on parallel cross-lingual corpora. To further refine entity alignment, we fine-tune LLMs on English-oriented entity-aligned cross-lingual corpora (\textsc{Eacl}), sourced from multilingual Wikipedia.
% To this point, we can complete ZCL-NER without parallel cross-lingual corpora. Additionally, to further enhance the alignment ability of the translation process,
% we employ the multilingual-supported Wikipedia to collect english-oriented entity-aligned cross-lingual corpora (EACL) to further optimize the LLMs' understanding of multilingual low-resource entities by fine-tuning, thereby enhancing the translation ability of entity alignment. 
In general, our contributions are summarized:

$\bullet$ We identify the linguistic script discrepancies and entity misalignment between LSL and NSL for ZCL-NER.

$\bullet$ We propose an entity-aligned translation (\textsc{Eat}) framework from the perspective of enhancing entity-aware translation ability for various NSL as the target languages.

$\bullet$ We introduce two metrics (BLEU and Shannon Entropy) to  quantify the correlation between translation quality and ZCL-NER performance.

\begin{figure*}[t]
  \centering
  \includegraphics[width=\textwidth]{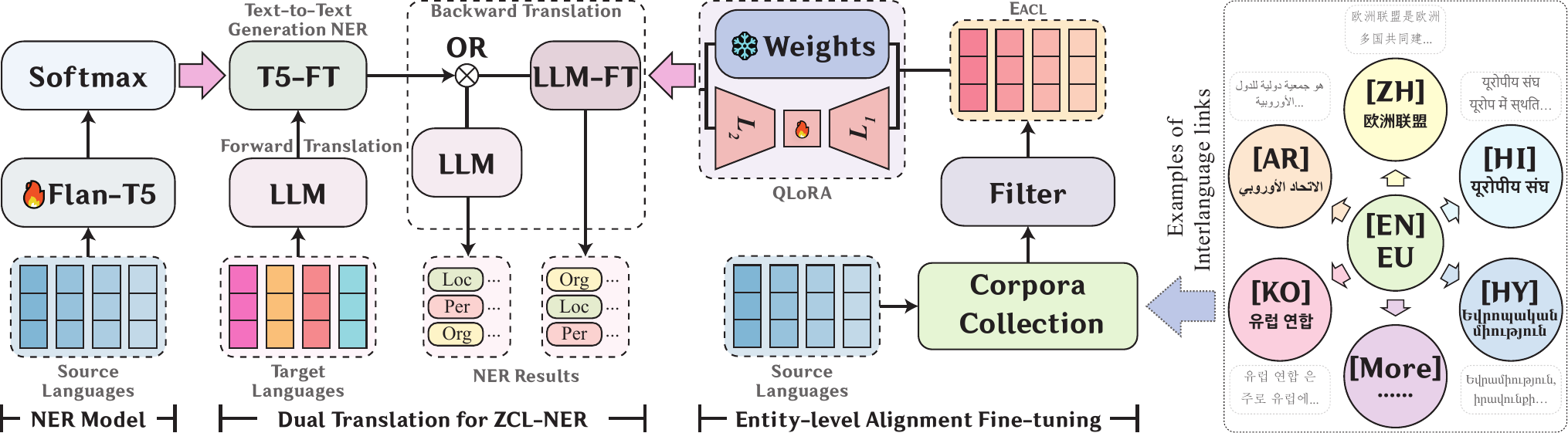}
  \caption{The overall architecture of our proposed \textsc{Eat} approach.}
  \label{fig:structure}
\end{figure*}

\section{Related Work}
\subsection{Linguistic Differences Between Scripts} Previous research has explored how linguistic and cognitive processes are influenced by different writing systems, such as Latin scripts, Hanzi, and Devanāgarī. \citet{GELMAN1998215} argue that formal language properties (e.g., scripts, morphemes) impact the use of generic noun phrases, which are essential for knowledge organization and reasoning~\cite{GELMAN1986183, FINIASZ2024105707}. The Script Relativity Hypothesis~\cite{pae2020script} suggests that script influences cognitive processes, with first-language experiences shaping how we process other languages~\cite{Li2022, Singh2022}. Systemic functional linguistics~\cite{halliday2006construing} also highlights how information is conveyed differently across languages due to their unique semantic and syntactic structures~\cite{YANG2008450, CHEN201671, Hita03042018}. In contrast to previous studies, which generalize NER models without considering linguistic typology, our approach leverages LLMs to bridge these gaps, yielding improved performance.

\subsection{Zero-shot Cross-Lingual NER} Recent zero-shot CL-NER approaches, particularly those using teacher-student (T-S) learning frameworks, have achieved promising results by distilling NER knowledge from source languages to target languages~\cite{Single-TS, RIKD, AdvPicker, DualNER}. Some works focus on improving knowledge distillation by reducing noise~\cite{MSD, ProKD, DenKD}. Additionally, machine translation is employed to generate pseudo-training data for CL-NER, using methods such as dictionary-based translation~\cite{cheap, neural}, sequence translation models~\cite{mulda, crop} and label projections~\cite{EasyProject, clap} for data augmentation. However, these methods often underperform compared to the T-S learning framework,
mainly because of their limited entity-aligned translation capabilities.
% likely due to their entity-aligned translation ability. 
While T-S frameworks achieve strong results for LSL, they struggle with NSL, such as Chinese and Japanese, where linguistic differences hinder accurate semantic and lexicogrammatical alignment.

\subsection{LLMs for NER and Multilinguality} Large Language Models (LLMs) have demonstrated significant potential across NLP tasks, including NER~\cite{xie2023emnlp, xie2024acl,cikm/LiuZLZC24}. LLMs have also been used for data augmentation to enhance the performance of smaller models~\cite{aaai/ZhangWLWZZ21,cdner, guidance,mm/JuZZLLZ24}. Recent studies have further improved their multilingual capabilities, enabling cross-lingual applications \cite{acmmm/multi-cul}. However, due to performance disparities between LLMs' English and non-English capabilities~\cite{q-trans-train}, many studies use LLMs to translate non-English texts into English before performing downstream tasks~\cite{q-trans, chen-etal-2024-breaking}. Inspired by these approaches, we utilize LLMs for entity-aligned translation, enhancing the alignment between source and target languages. The advanced semantic understanding and reasoning abilities of LLMs help capture implicit information during translation, mitigating information loss and improving overall performance.

\section{Methodology}
In this section, we first introduce our proposed \textbf{E}ntity-\textbf{A}ligned \textbf{T}ranslation (\textbf{\textsc{Eat}}) framework, as illustrated in Figure \ref{fig:structure}: dual translation, source-oriented cross-lingual corpora collection, and entity-level alignment fine-tuning. Then, we present two metrics to evaluate the correlation between translation quality and ZCL-NER performance.

% \textbf{Task Formulation.} Given an $n$-token sentence $\boldsymbol{x} = <x_1,\cdots,x_n>$ and $k$-type entity set $\bm\tau = <t_1,\cdots,t_k>$, the object of NER task is to extract all entities $\boldsymbol{e}_i \in \boldsymbol{E}$ from $\boldsymbol{x}$ and assign one of the types in $\bm\tau$ to each entity, where $\boldsymbol{e}_i = (\boldsymbol{x}_{start:end}, t)$ denotes the $i$-th entity of $\boldsymbol{x}$ and $t \in \bm{\tau}$ refers to the type of the entity. $\boldsymbol{x}_{start:end}$ refers to a continues word span $<x_{start},\cdots,x_{end}>$ in $\boldsymbol{x}$, where $start$ and $end$ refers to the entity boundary indexes respectively. Given dataset $\mathcal{D}_s$ of the source language (i.e., English in our setting) and dataset $\mathcal{D}_t$ of the target language, the objective of the ZCL-NER task is to acquire target-related knowledge from $\mathcal{D}_s$ to enhance model's performance on $\mathcal{D}_t$.

\subsection{Dual Translation for ZCL-NER}
% Although it is a common practice to leverage LLMs to directly translate sentences from one language to another, there are several problems that could interfere with the result of translation. The most significant problem is hallucination. It refers to LLMs sometimes not being able to follow the instructions, providing some answers irrelevant to targeted tasks, and even fabricating seemingly correct answers.
To minimize entity loss during translation, we propose a dual translation model (\textbf{DT}) that focuses on preserving potential entities of target language throughout the process.

\textbf{Target to Source Forward Translation.}
% (target language to English) 
LLMs may not be able to paraphrase the words of entities in single round of inference and may directly output the raw input words.
% In the case of translating languages that use Latin scripts, this defect may not have significant impacts since those languages already have shared words, and the absence of some words does not affect the comprehension of the whole sentence.
Therefore, the generated translation with raw words from languages that do not use Latin scripts will affect the models' overall understanding and may lead to failure in subsequent tasks. One possible solution is to instruct the LLMs in several rounds instead of single round in each direction, which we call \emph{multi-round text translation with chain-of-thought} (MrCoT).

In the first round, we instruct the model to \textbf{consider the entities the target sentence $\boldsymbol{x}$ may contain} and explain those entities, when translating from target language $a$ to source language (English) $b$. Formally,
\begin{equation}
    o_1^t = \textsc{Lm}(p_1^t, \boldsymbol{x}, a, b)
\end{equation}
where $\textsc{Lm}$ denotes a large language model. $p_1^t$ is the first prompt that instructs the model to consider the entities where $\boldsymbol{x}$ may contain and describe them. $o_1^t$ is supposed to provide a CoT context for the next round.
Then the second round output is:
\begin{equation}
    o_2^t = \textsc{Lm}^{o_1^t}(p_2^t, \boldsymbol{x}, a, b)
\end{equation}
where $p_2^t$ is the second prompt that instructs the model to translate $\boldsymbol{x}$ taking previous inference $o_1^t$ into consideration. The translation result $\mathcal{T}_{a\rightarrow{b}}^t(\boldsymbol{x})$ is obtained by filtering non-relevant words:
\begin{equation} \mathcal{T}_{a\rightarrow{b}}^t(\boldsymbol{x}) = \textsc{Lm}^{o_2^t}(p^f,\boldsymbol{x},a,b)
  % \mathcal{T}_{a\rightarrow{b}}^t(\boldsymbol{x}) = \textsc{Lm}({F}, o_2^t)
\end{equation}
where $p^f$ is the filter prompt. $\mathcal{T}_{a\rightarrow{b}}^t(\boldsymbol{x})$ denotes the complete sentence by target-to-English language translation.

% 1) In the first round, we instruct the model to translate $\boldsymbol{x}$ from $a$ to $b$ while considering the entities it may contain.
% The first-round results contain the translation with the model's thinking process.
% This round is supposed to provide a CoT context for the next round.
% 2) Then, we instruct the model to translate the sentence taking previous inference into consideration to obtain more accurate translations.
% 3) Finally, we ask the model to output the translated sentence as a result.

\textbf{Text-to-Text Generation for NER.} To better utilize the semantics of the sentences, we reformulate the NER task as a text-to-text generation task following \cite{cdner}. The inputs are divided as: 1) \textbf{PREFIX}($P$): define the task as labeling entities of the input sentence. 2) \textbf{TAG}: the set $\mathbb{T}$ of entity tags from the dataset. 3) \textbf{SENTENCE}: the input sentence $\mathcal{T}^t_{a\rightarrow{b}}(\boldsymbol{x})$. Then, given the entire input $I = ({P},\mathbb{T},\mathcal{T}^t_{a\rightarrow{b}}(\boldsymbol{x}))$,
% \begin{equation}
%     I = ({P},\boldsymbol{t},\boldsymbol{x})
% \end{equation}
the output entities by generation model are defined as:
\begin{equation}
    \boldsymbol{E} = \textsc{Extractor}_{\boldsymbol{\rho}}(I)
\end{equation}
where $\boldsymbol{\rho}$ denotes the trainable parameters of the text-to-text English NER model $\textsc{Extractor}$. 
% The final result $\boldsymbol{E}$ is obtained:
% \begin{equation}
%     \boldsymbol{E} = \textit{Softmax} ( \mathcal{L} (\boldsymbol{y}) )
% \end{equation}
% where $\mathcal{L}$ denotes the linear layer.
The result $\boldsymbol{E}$ contains entity pairs as $(\mathcal{T}^t_{a\rightarrow{b}}(\boldsymbol{x})_{l_1:r_1}, tag)$, where $l_1:r_1$ denotes the text boundary.

\textbf{Source to Target Backward Translation.} This stage plays a crucial role in ensuring entity alignment across languages. In neural machine translation, LLMs may generate tokens that are semantically related to entities, rather than directly translating them. As a result, relying solely on the LLM-generated translation may lead to the omission of possible entities that were captured in the above stage. To ensure effective cross-lingual entity alignment, we check whether the translated results of this stage for potential input entities in English \textbf{(i.e., output) closely match the corresponding segments in the real target language}.
% In neural machine translation, LLMs may generate tokens related to entities, rather than directly translating them. Therefore, simply using the translation of LLMs will still lead to the missing of potential entities that were previously captured. Considering this, for cross-lingual entity alignment, we hope that the result fragments of the current back-translation should appear in the target language of the previous stage as much as possible.

% \textbf{Entity-level Translation Alignment}
% (English entity to target language)
% In neural machine translation, the LLMs may generate tokens that are related to the entities rather than translating them. In addition, one focus of the NER tasks is recall, which means that the generated translation tokens should be segments of the origin inputs. In other words, the NER results should align to the input sentences. Here we use similar methods to translate the entities and align them to the raw inputs.

Formally, to translate an entity $\mathcal{T}_{a\rightarrow{b}}^t(\boldsymbol{x})_{l_1:r_1}$ from language $b$ to $a$, the first round output is:
\begin{equation}
    o_1^e = \textsc{Lm}(p_1^e, \mathcal{T}_{a\rightarrow{b}}^t(\boldsymbol{x})_{l_1:r_1}, \boldsymbol{x}, a, b)
\end{equation}
where $l_1:r_1$ is the text boundary, and $p_1^e$ is the first prompt that instructs the model to translate the entity and analyze if $o_1^e$ could appear in $\boldsymbol{x}$. Then the translation result in second round with MrCoT is obtained similarly:
\begin{equation}
    o_2^e = \textsc{Lm}(p_2^e, o_1^e, \boldsymbol{x}, a, b)
\end{equation}
\begin{equation}
    \boldsymbol{x}_{l_2:r_2} = \mathcal{T}_{b\rightarrow{a}}^e(\mathcal{T}_{a\rightarrow{b}}^t(\boldsymbol{x})_{l_1:r_1}) = \textsc{Lm}(p^f, o_2^e)
\end{equation}
where $l_2:r_2$ is the text boundary, and $p_2^e$ is the second prompt to instruct the model to check if the result appears in $\boldsymbol{x}$, and $p^f$ is the filter prompt.

At this stage, we can finalize ZCL-NER, where $\boldsymbol{x}_{l_2:r_2}$ represents \textit{the target language entity we aim to extract}.
% At this point, we can complete ZCL-NER already, in which $\boldsymbol{x}_{l_2:r_2}$ is \textit{just the entity of target language we want.}

% 1) In the first round, we ask the LLM to translate the English entity into the target language while the result should appear in the given sentence.
% 2) Then we instruct the LLM to check if the result appears in the given sentence.
% 3) Finally, we ask the model to output the translated entity as results.

\subsection{Source-oriented Cross-lingual Corpora}
% \subsection{Source-oriented Cross-lingual Corpora Collection}
\label{sec:et-corpus}
To reduce the hallucination and further amplify the ability of entity alignment in the above translation process, we collect English-oriented entity-aligned cross-lingual corpora ({\textsc{Eacl}}) to fine-tune the translation model.
The \textsc{Eacl} Corpora are collected from \textbf{Interlanguage Links}~\protect\footnotemark[1] provided by the {Wikipedia API}~\protect\footnotemark[2].

\footnotetext[1]{Detailed information in \url{https://en.wikipedia.org/wiki/Help:Interlanguage_links}}
\footnotetext[2]{\url{https://api.wikimedia.org/wiki/Main_Page}}

% We leverage the entities from the CoNLL2003 English dataset to construct the E-T Corpora. As shown in Figure~\ref{fig:structure}, for an English entity, we use the Wikipedia API to fetch all the corresponding pages in other languages (namely Interlanguage Links). Each corresponding page contains a title and a short summary of the entity in another language, where the title and summary can be regarded as the entity and the text in which it appears.

We leverage the entities from the CoNLL2003 English dataset to construct \textsc{Eacl}. As shown in Figure~\ref{fig:structure}, for an English entity $e \in \boldsymbol{e}$, the links from the API of $e$ are the content of target language related to $e$:
\begin{equation}
    \mathcal{I}(e) = \left\{ ({u}^{a}, \boldsymbol{v}^a)_e | a \in \mathcal{A} \right\}
\end{equation}
where ${u}^a$ and $\boldsymbol{v}^a$ denote the title (entity or short description for $e$ in target language $a$) and summaries of language $a$ corresponding to $e$ (a text to explain and describe title), and $\mathcal{A}$ denotes the set of languages. We leverage $u^a$ as the entity, and the first sentence $v_1^a$ of $\boldsymbol{v}^a$ as the text to construct the entity-description pair: $\mathcal{D}^a = \left\{ \mathcal{I}_a(e) | e \in \boldsymbol{e} \right\} = \left\{ ({u}^{a}, {v}^a_1)_e | e \in \boldsymbol{e} \right\}$.
% The dataset of language $a$ is:
% \begin{equation}
%     \mathcal{D}^a = \left\{ \mathcal{I}_a(e) | e \in \boldsymbol{e} \right\} = \left\{ ({u}^{a}, {v}^a_1)_e | e \in \boldsymbol{e} \right\}
% \end{equation}

However, not all entities have corresponding Wikipedia pages with interlanguage links. Moreover, not all languages have Wikipedia pages that align with English entities, especially in less-resourced languages. As a result, the corpus size varies across languages. Detailed information on the collected corpora can be found in Table~\ref{tab:etcor-detail}.

\begin{table}[t]
\centering
\small
% \resizebox{\linewidth}{!}{
  \begin{tabular}{l|cccc}
     \toprule
     {} & {\textbf{AR}} & {\textbf{HI}} & {\textbf{HY}} & {\textbf{JA}} \\
     \midrule
     Sen. & \ \ 2,746 & \ \ 1,195 & \ \ 1,568 & \ \ \ \ 2,890 \\
     Tok. & 69,336 & 83,287 & 62,669 & 217,252 \\
     \midrule
     {} & {\textbf{KA}} & {\textbf{KO}} & {\textbf{RU}} & {\textbf{ZH}} \\
     \midrule
     Sen. & 1,451 & \ \ 2,014 & \ \ 796 & \ \ \ \ 2,382 \\
     Tok. & 8,177 & 32,667 & 7,521 & 206,279 \\
     \bottomrule
  \end{tabular}
  % }
  \caption{Detailed information of our collected \textsc{Eacl} Corpora, including the amounts of $({u}^{a}, {v}_1^a)_e$ and tokens in ${v}_1^a$ for each language $a$.}
  \label{tab:etcor-detail}
\end{table}

\subsection{Entity-level Alignment Fine-tuning}
\label{sec:sft}
To amplify the DT model's entity-level alignment ability, we leverage \textsc{Eacl} Corpora obtained above to fine-tune (FT) the backward translation. This is because the model struggles to identify the positions of entities.
Specifically, we leverage Quantized-LoRA (QLoRA)~\cite{QLORA} to accelerate model fine-tuning under constrained resources.
Instead of global quantization, we use block-wise $k$-bit quantization~\cite{iclr/DettmersLSZ22,acl/TangLWZLCZ25}:
% for its improvements in terms of accuracy, efficiency, and flexibility. For a single layer, the parameters of the model with LoRA adapter: 
% \begin{equation}
%     \boldsymbol{Y} = \boldsymbol{X}{D}^2 (\boldsymbol{c}, \boldsymbol{c}^d, \boldsymbol{W}^q) + \boldsymbol{X}\boldsymbol{L}_1\boldsymbol{L}_2
% \end{equation}
\begin{equation}
    {D}^2(\boldsymbol{c}, \boldsymbol{c}^d, \boldsymbol{W}^q) = D(D(\boldsymbol{c}, \boldsymbol{c}^d), \boldsymbol{W}^q) = \boldsymbol{W}
\end{equation}
where the quantization constants $\boldsymbol{c}$ are quantized as $\boldsymbol{c}^d$. $\boldsymbol{W}$ and $\boldsymbol{W^q}$ denote the model's raw and quantized weights.
% , and $\boldsymbol{L}_1\boldsymbol{L}_2$ denotes the trainable parameters of the LoRA adapters.

% Besides, cross-entropy loss is optimized to train the DT model:
% \begin{equation}
%   L_T = -\frac{1}{\eta}\sum_{j=1}^{\eta} \log x_j y_j
% \end{equation}
% where $x_j \in \boldsymbol{x}$ denotes the input template which is constructed using the entity-description pair. $y_j \in \boldsymbol{y}$ denotes the output of the model, and $\eta$ denotes the max input sequence length of the model.

We employ cross-entropy loss to train the DT model:
% \begin{equation}
%     \boldsymbol{y} = \textsc{Lm}(e, {v}_1^a)
% \end{equation}
\begin{equation}
    L_T(\boldsymbol{y},\boldsymbol{\hat{y}}) = -\sum_{i=1}^{\eta}\hat{y}_i\log (y_i)
\end{equation}
where $\boldsymbol{y} = \textsc{Lm}(e, {v}_1^a)$. $\hat{y}_i \in \boldsymbol{\hat{y}}$ is $u^a$ in $(u^a,v_1^a)_e$ of \textsc{Eacl}. $y_i \in \boldsymbol{y}$ denotes the predicted entity. $\textsc{Lm}$ denotes a large language model and $\eta$ denotes the max length of model output.

% \subsection{Text-to-text Generation for NER}
% To better utilize the semantics of the sentences, we reformulate the NER task as a text-to-text generation task following \cite{emnlp/ZhangLWZLCZ24}. The inputs are divided as: 1) \textbf{PREFIX}($P$): define the task as labeling entities of the input sentence. 2) \textbf{TAG}: the entity tags $\boldsymbol{t}$ from the dataset. 3) \textbf{SENTENCE}: the input sentence $\mathcal{T}^t_{a\rightarrow{b}}(\boldsymbol{x})$. Then the entire input is $I = ({P},\boldsymbol{t},\mathcal{T}^t_{a\rightarrow{b}}(\boldsymbol{x}))$.
% % \begin{equation}
% %     I = ({P},\boldsymbol{t},\boldsymbol{x})
% % \end{equation}
% The model takes $I$ as input and outputs the generation $\boldsymbol{y}$ which contains the entities:
% \begin{equation}
%     \boldsymbol{y} = \textsc{Lm}_{\boldsymbol{\rho}}(I)
% \end{equation}
% where $\boldsymbol{\rho}$ denotes the trainable parameters of the model \textsc{Lm}. The final result $\boldsymbol{E}$ is obtained:
% \begin{equation}
%     \boldsymbol{E} = Softmax ( \mathcal{L} (\boldsymbol{y}) )
% \end{equation}
% where $\mathcal{L}$ denotes the linear layer, and the result $\boldsymbol{E}$ contains entity pairs as $(\mathcal{T}^t_{a\rightarrow{b}}(\boldsymbol{x})_{l_1:r_1}, tag)$. 

% Figure~\ref{fig:ner} shows an example of the NER progress described above.

% \begin{figure}[t]
%   \centering  
%   \includegraphics[width=\linewidth]{ner-bold.pdf}
%   \caption{The simple structure of text-to-text generation with instructor.}
%   \label{fig:ner}
% \end{figure}

% \subsection{Entity-aligned Translation Ability Evaluation}
\subsection{Evaluating Entity-aligned Translation}
\label{sec:bleu}
We use BLEU score~\cite{bleu} and Information Entropy~\cite{ie} to measure the information loss in the translation process, so as to demonstrate the relevance between the NER results and the information loss.

\textbf{Bilingual Evaluation Understudy} (BLEU) is widely used as an evaluation metric in machine translation due to its fast and unified features. In our setting, the BLEU score is evaluated between the generated target sentence in backward translation and original input in forward translation. 
% Formally, it is finally calculated by the geometric mean of the $n$-gram precision:
% \begin{equation}
%     \mathop{BLEU} = \textit{BP}\times \exp (\frac{1}{4}\sum_{n=1}^4 \log(\mathcal{P}_n))
% \end{equation}
% where $\textit{BP}$ denotes Brevity Penalty and $\mathcal{P}_n$ represents $n$-gram precision \cite{bleu}.

\textbf{Information Entropy} (Shannon Entropy) is also commonly used to quantify the information of the sentences. For better analysis, we leverage the Bi-Gram Model to calculate the joint information entropy as: $H(\boldsymbol{s}) = \sum_{i=1}^{n}H(s_{i-1}, s_{i})$.
% \begin{align}
%     \begin{aligned}
%     H(\boldsymbol{s}) &= \sum_{i=1}^{n}H(s_{i-1}, s_{i}) \\
%                       &= \sum_{i=1}^{n} P(s_{i-1}, s_{i}) (-\log P(s_{i-1}|s_{i}))
%     \end{aligned}
% \end{align}
% where $P(s_{i-1}, s_{i})$ denotes the joint probability of $s_{i-1}$, $s_{i}$ appearing in the $n$-length text $\boldsymbol{s}$ with $s_{i-1}$ exactly before $s_{i}$, and $P(s_{i-1}|s_{i})$ denotes the conditional probability of $s_{i-1}$ appearing before $s_{i}$.

To evaluate the information loss between raw target sentence and dual translated sentence, we define the measurement using the above Shannon Entropy as: $L_e = \frac{H(\mathcal{T}_{b\rightarrow{a}}^t(\boldsymbol{s}_r))}{H(\boldsymbol{s}_r)}$,
% \begin{equation}
%     L_e = \frac{H(\mathcal{T}_{b\rightarrow{a}}^t(\boldsymbol{s}_r))}{H(\boldsymbol{s}_r)}
% \end{equation}
where $\boldsymbol{s}_r$ denotes the sentence of target language $b$ and $a$ denotes the source language.

\begin{table}[t]
\centering
\small
% \resizebox{\linewidth}{!}{
  \begin{tabular}{l|cccc}
     \toprule
     \multirow{2}*{\textbf{Lang.}} & \multicolumn{3}{c}{Tokens} & {\multirow{2}*{Ratio}} \\
      & Train & Valid & Test & \\
     \midrule
     AR & 129,184 & \ \ 64,291 & \ \ 64,347 & \ \ 4.96\% \\
     HI & \ \ 29,443 & \ \ \ \ 5,808 & \ \ \ \ 6,005 & 11.05\% \\
     HY & \ \ 95,614 & \ \ \ \ 6,214 & \ \ \ \ 6,220 & \ \ 0.08\% \\
     JA & 603,301 & 300,844 & 306,959 & \ \ 1.60\% \\
     KA & \ \ 80,402 & \ \ 81,159 & \ \ 81,922 & \ \ 0.05\% \\
     KO & 162,031 & \ \ 80,786 & \ \ 80,841 & \ \ 1.06\% \\
     RU & 141,529 & \ \ 70,279 & \ \ 71,288 & \ \ 3.33\% \\
     ZH & 420,054 & 213,682 & 207,505 & 17.75\% \\
     % DE & 195,387 & \ \ 97,805 & \ \ 97,646 & \ \ 1.75\% \\
     % ES & 129,283 & \ \ 64,329 & \ \ 64,728 & \ \ 7.30\% \\
     % FR & 136,788 & \ \ 68,220 & \ \ 68,754 & \ \ 4.07\% \\
     % NL & 169,449 & \ \ 84,146 & \ \ 85,122 & \ \ 0.39\% \\
     \midrule
     EN & 160,394 & \ \ 80,536 & \ \ 80,326 & 19.76\% \\
     \bottomrule
  \end{tabular}
  % }
  \caption{Detailed information of our selected languages in the dataset WikiANN. Ratio represents the approximate proportion of speakers to the total world population, and the statistics are referred from Wikipedia. }
  \label{tab:cor-detail}
\end{table}

\begin{table*}[t]
  \centering
  \small
  \begin{tabular}{l|cccccccc|c}
     \toprule
     \multirow{2}*{\textbf{Models}} & \multicolumn{8}{c}{\textbf{Non-Latin Scripts}} \\
          & \textbf{AR} & \textbf{HI} & \textbf{HY} & \textbf{JA} & \textbf{KA} & \textbf{KO} & \textbf{RU} & \textbf{ZH} & \textbf{Avg.}\\
     \midrule
     mBert~\cite{FTDT}          & 42.30 & 64.79 & 52.12 & 29.82 & 64.68 & 57.38 & 64.09 & 43.85 & 52.38 \\
     A-align\cite{Awesome-align} & 46.00 & 73.90 & 49.83 & 20.30 & 70.40 & 57.70 & 64.80 & 45.40 & 53.54 \\
     CROP~\cite{crop}  & 52.44 & 55.55 & 44.49 & 45.37 & 46.02 & 48.93 & 50.73 & 45.33 & 48.61 \\
     EasyProject\cite{EasyProject} & 34.40 & 73.00 & 48.61 & 41.30 & 66.40 & 48.20 & 66.30 & 42.00 & 52.54 \\
     CLaP\cite{clap} & 48.70 & 73.10 & 51.68 & 45.30 & 70.50 & 60.10 & 68.30 & 49.70 & 58.42 \\
     \midrule
     TSLM~\cite{Single-TS}      & 43.12 & 65.26 & 53.56 & 31.19 & 66.20 & 58.94 & 66.02 & 45.60 & 53.74 \\
     RIKD~\cite{RIKD}           & 45.96 & 65.69 & 55.17 & 31.49 & 66.83 & 58.03 & 65.63 & 47.38 & 54.52 \\
     AdvPicker~\cite{AdvPicker} & 49.16 & 70.00 & 52.49 & 37.62 & 68.37 & 59.25 & 68.28 & 53.02 & 57.27 \\
     % MulDA~\cite{mulda}         & 53.62 & 67.46 &       & 37.05 & 67.68 & 52.69 & 65.00 & 41.77 &       \\
     DualNER~\cite{DualNER}     & 59.00 & 66.24 & 55.92 & 31.07 & 67.28 & 57.48 & 65.06 & 47.84 & 56.24 \\
     MSD~\cite{MSD}             & 62.88 & 73.43 & 56.22 & 33.34 & 69.23 & 61.44 & 67.71 & 57.06 & 60.16 \\
     ProKD~\cite{ProKD}         & 50.91 & 70.72 & 62.58 & 33.72 & 69.07 & 61.31 & 65.59 & 51.80 & 58.21 \\
     DenKD~\cite{DenKD}         & 60.01 & 69.76 & 65.20 & 37.90 & 69.30 & 62.51 & \underline{69.35} & 55.62 & 61.21 \\
     \midrule
     \textsc{Eat} w/o FT & \underline{66.53} & \textbf{76.26} & \underline{65.62} & \underline{45.43} & \underline{71.63} & \textbf{66.03} & \textbf{71.45} & \underline{60.12} & \underline{65.38} \\
     \textsc{Eat} & \textbf{67.29} & \underline{75.46} & \textbf{68.02} & \textbf{52.26} & \textbf{73.68} & \underline{65.23} & 63.25 & \textbf{61.27} & \textbf{65.81} \\
     \bottomrule
  \end{tabular}
  \caption{Performance comparison of existing zero-shot CL-NER studies and our approaches. Bold represents the best result, and underlining represents the second best result.}
  \label{tab:main}
\end{table*}

\section{Experimentation}
\label{sec:expe}
\subsection{Datasets}
The experiments are conducted on two public and most widely used datasets, including WikiANN~\cite{wikiann} and CoNLL2003~\cite{conll03}:

1) \textbf{WikiANN} involves 176 languages, and each language has balanced train, valid, and test splits. The entity categories of WikiANN are PERSON (PER), LOCATION (LOC), and ORGANIZATION (ORG). We select 8 non-Latin script languages as our test sets (target language), representing both widely and less widely spoken languages: Arabic (AR), Hindi (HI), Armenian (HY), Japanese (JA), Georgian (KA), Korean (KO), Russian (RU), and Chinese (ZH).
% In addition, we also select 4 Latin script languages for comparison, including German (DE), Spanish (ES), French (FR), and Dutch (NL).
The English (EN) train and valid sets (source language) are leveraged to build the NER Extractor. Detailed information and statistics about these languages are presented in Table~\ref{tab:cor-detail}.

2) \textbf{CoNLL2003} includes two languages: English and German. Each language has train, valid, and test splits. Due to its higher quality of manual annotation compared to WikiANN, we use the entity phrases only from the English train set to construct the \textsc{Eacl} Corpora.

% , as described in Section~\ref{sec:et-corpus}.

% \subsection{Implementation Details}
 
% The BLEU scores and the entropy loss are calculated as described in Section~\ref{sec:bleu}.

% \begin{table}[ht]
% \centering
% \small
% \resizebox{\linewidth}{!}{
%   \begin{tabular}{l|cccc|cc}
%      \toprule
%      {} & {\textbf{KO}} & {\textbf{JA}} & {\textbf{ZH}} & {\textbf{AR}} & {\textbf{FR}} & {\textbf{ES}} \\
%      \midrule
%      Sen. & 2014 & 2890 & 2382 & 2746 & 2331 & 2197\\
%      Tok. & 32667 & 217252 & 206279 & 69336 & 61887 & 67792\\
%      \midrule
%      {} & {\textbf{RU}} & {\textbf{KA}} & {\textbf{HY}} & {\textbf{HI}} & {\textbf{DE}} & {\textbf{NL}} \\
%      \midrule
%      Sen. & 796 & 1451 & 1568 & 1195 & 2335 & 1797\\
%      Tok. & 7521 & 8177 & 62669 & 83287 & 35902 & 34450\\
%      \bottomrule
%   \end{tabular}
%   }
%   \caption{Detailed information of our collected Entity-translation Corpora, including the amounts of entity-sentence pairs and tokens.}
%   \label{tab:etcor-detail}
% \end{table}

\subsection{Baselines and Implementation}
\textbf{Baselines.} To compare our approach with previous translation-based and teacher-student approaches, we report the baselines as follows:

Translation based: 1) \textbf{mBert}~\cite{FTDT}. 2) \textbf{A-align}~\cite{Awesome-align} 3) \textbf{CROP}~\cite{crop} 4) \textbf{EasyProject}~\cite{EasyProject} 5) \textbf{CLaP}~\cite{clap}.

T-S based: 3) \textbf{TSLM}~\cite{Single-TS}. 4) \textbf{RIKD}~\cite{RIKD}.
5) \textbf{AdvPicker}~\cite{AdvPicker}.
6) \textbf{DualNER}~\cite{DualNER}.
7) \textbf{MSD}~\cite{MSD}.
8) \textbf{ProKD}~\cite{ProKD}.
9) \textbf{DenKD}~\cite{DenKD}, the \emph{SOTA} for ZCL-NER.

\textbf{Implementation Details.} The \textsc{Eacl} Corpora are leveraged to fine-tune the model \textbf{Qwen2.5-14B-Instruct}~\cite{qwen2, qwen2.5}. We first quantize the model to $4$-bits, and freeze its parameters. The rank parameter and the scale parameter of the Low-Rank Adapter are set to $64$ and $16$. The ratio of train and valid sets is $90$: $10$, and the model is trained using the divided corpora for $5$ epochs with the learning rate set to ${1.0e{-4}}$.

We leverage English train and valid sets from the WikiANN dataset to train the \textbf{Flan-T5-Base} model~\cite{flant5}. After $50$ epochs of training with a learning rate of ${1.0e{-4}}$, we will obtain the final NER Extractor.

Since there are no reference translation results (entities) for the WikiANN dataset, we utilize the raw input $\boldsymbol{x}$ of target language and dual translated text $\mathcal{T}_{b\rightarrow{a}}^t(\mathcal{T}_{a\rightarrow{b}}^t(\boldsymbol{x}))$ to calculate the \textbf{BLEU} scores and \textbf{Entropy Loss}. 
% Specifically, the target language sentences are translated to the source language (English). Then the results are translated back into the target language. We use the raw sentences as the references and the twice-translated sentences as the candidates.

% \begin{table}[t]
% \centering
% \small
% % \resizebox{\linewidth}{!}{
%   \begin{tabular}{l|cccc}
%      \toprule
%      {} & {\textbf{AR}} & {\textbf{HI}} & {\textbf{HY}} & {\textbf{JA}} \\
%      \midrule
%      Sen. & \ \ 2,746 & \ \ 1,195 & \ \ 1,568 & \ \ \ \ 2,890 \\
%      Tok. & 69,336 & 83,287 & 62,669 & 217,252 \\
%      \midrule
%      {} & {\textbf{KA}} & {\textbf{KO}} & {\textbf{RU}} & {\textbf{ZH}} \\
%      \midrule
%      Sen. & 1,451 & \ \ 2,014 & \ \ 796 & \ \ \ \ 2,382 \\
%      Tok. & 8,177 & 32,667 & 7,521 & 206,279 \\
%      \bottomrule
%   \end{tabular}
%   % }
%   \caption{Detailed information of our collected EACL Corpora, including the amounts of entity-text pairs and tokens (exclude the ShareGPT template's tokens).}
%   \label{tab:etcor-detail}
% \end{table}

\subsection{Main Results}
We use the token-level micro F1 score to evaluate the NER results, following previous works~\cite{ProKD, DenKD, MSD}. The baseline results of T-S based approaches are cited from their papers. 
% To better compare our proposed approach, we conduct our \ with and without supervised fine-tuning (SFT). 
Based on the performance comparison in Table~\ref{tab:main}, we primarily address the following questions:

\textbf{How does \textsc{Eat} perform and why do we design it?}
Translation-based methods perform poorly in NSL. This indicates that the traditional translation schema between English and NSL is not suitable for ZCL-NER. Therefore, we need to develop a completely new translation mechanism to account for the significant differences between English and NSL, which aligns with our core motivation. Furthermore, although T-S framework (e.g., DenKD) apparently outperforms previous translation methods (e.g., CLaP), the principles of the T-S series remain largely similar. As a result, its effectiveness in bridging the linguistic gap between English and NSL is limited. Consequently, performance gains over previous works from the past two years have remained modest, averaging only at 1\%-2\%. However, our \textsc{Eat} achieves a substantial improvement of 4\% over SOTA. This suggests that our approach genuinely addresses the fundamental disparities between different languages.

% our approach achieves better performance on most languages than the SOTA approach DenKD with an improvement on the average F1 by $4.60\%$ with FT and $4.17\%$ without FT. 
% To be specific, 
% Our \textsc{Eat} outperforms previous best results by $1.92\%$ to $4.41\%$ for \textbf{fusional} languages, $3.52\%$ to $6.89\%$ for \textbf{agglutinative} languages, and $4.21\%$ for \textbf{isolating} languages. In particular, compared to the teacher-student approaches, our approach achieves a huge boost by $14.36\%$ in Japanese (JA).

% \textbf{How does \textsc{Eat} perform on specific language?}
In particular, the T-S based approaches perform particularly poorly on JA and ZH, almost at 30\%-55\%. However, we found that there are a significant number of people worldwide speak these languages, as the ratio in Table \ref{tab:cor-detail}. This motivates us to enhance the performance of NER for these languages. Therefore, we develop an intuitive and effective approach that significantly outperforms SOTA in ZCL-NER.

% \textbf{Is it necessary to collect EACL for fine-tuning?}
% Although the average performance has only improved by 0.5\% after using fine adjustment, the performance improvement has been significantly improved in several individual languages. For example, there are 2-3 points for HY and KA language improvement. I found that the data volume of these two languages is the smallest, which means they are relatively scarce languages. This shows that in relatively scarce languages, we adopt additional data fine translation models, which is more likely to bring NER performance improvements.

\textbf{Why doesn't \textsc{Eat} w/ FT always perform better than it w/o FT?}
Although our \textsc{Eat} w/ FT performs well on average, it does not take effect in certain languages. One possible reason is inductive bias. The inductive bias is the structure imposed in the FT datasets to instruct the LLMs how to think, and it may be toxic:
\begin{quote}
    \emph{
    \protect\footnotemark[1]Clever structures posed by human researchers typically become the bottleneck when scaled up.
    }
\end{quote}
\footnotetext[1]{Hyung Won Chung (OpenAI). 2024-05. Don’t teach. Incentivize: Scale-first view of Large Language Models. MIT EI seminar.}

The fine-tuning on \textsc{Eacl} allows LLMs to learn new knowledge, but the structures introduced by FT may hinder LLMs' inference ability~\cite{kimiteam2025kimik15scalingreinforcement, deepseekai2025deepseekr1incentivizingreasoningcapability}. In other words, FT may force LLMs to think structurally, hence hindering the inference process where LLMs can correct previously generated errors. 

\subsection{Analysis and Discussion}

\begin{table}[t]
\centering
\small
\resizebox{\linewidth}{!}{
  \begin{tabular}{l|cccccccc}
     \toprule
     \multirow{2}{*}{} & \multicolumn{2}{c}{\textbf{AR}} & \multicolumn{2}{c}{\textbf{HI}} & \multicolumn{2}{c}{\textbf{JA}} & \multicolumn{2}{c}{\textbf{ZH}} \\
     & F1.$\uparrow$ & BLEU$\uparrow$ & F1.$\uparrow$ & BLEU$\uparrow$ & F1.$\uparrow$ & BLEU$\uparrow$ & F1.$\uparrow$ & BLEU$\uparrow$ \\
     \midrule
     \textsc{Eat}-14B    & {66.53} & 12.10 & 76.26 & 10.66 & 45.43 & 1.86 & {60.12} & 9.26 \\
     \textsc{Eat}-7B     & 65.30 & \ \ 9.70 & 73.89 & \ \ 8.57 & 40.25 & 1.60 & 59.73 & 8.81 \\
     \textsc{Eat}-3B     & 54.93 & \ \ 6.63 & 67.87 & \ \ 6.03 & 34.25 & 1.46 & 55.62 & 7.18 \\
     \textsc{Eat}-1.5B   & 46.51 & \ \ 4.10 & 60.59 & \ \ 2.22 & 28.06 & 0.57 & 48.46 & 5.54 \\
     \bottomrule
  \end{tabular}
}
  \caption{Performance comparison of our \textsc{Eat} w/o FT using homologous LLMs with different sizes of parameters.}
  \label{tab:ablation}
\end{table}

\begin{table}[t]
\centering
\small
% \resizebox{\linewidth}{!}{
  \begin{tabular}{l|cccc}
     \toprule
     {} & {\textbf{AR}} & {\textbf{HI}} & {\textbf{JA}} & {\textbf{ZH}} \\
     \midrule
     \textsc{Eat}-T5-NER    & {67.29} & 76.26 & 52.26 & {61.27} \\
     \textsc{Eat}-mBERT-NER  & 67.63 & 73.75 & 52.72 & 60.99 \\
     \bottomrule
  \end{tabular}
% }
  \caption{Performance comparison of our approach using different NER $\textsc{Extractor}$ models. LLM for DT keeps 14B for fair comparison.}
  \label{tab:ablation-ner}
\end{table}

\begin{figure}[t]
    \centering
\includegraphics[width=0.7\linewidth]{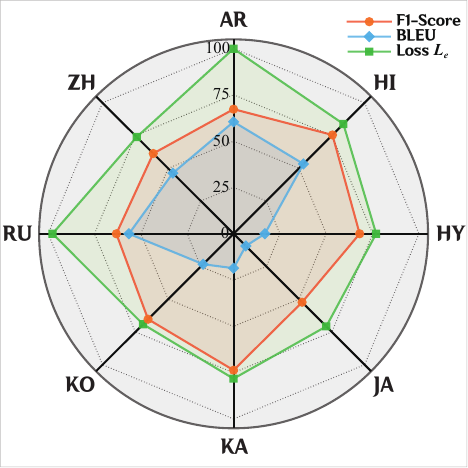}
    \caption{The relevance of BLEU scores and entropy loss compared with NER results (F1-scores). The BLEU scores are not the accurate values, we resize them to draw the plot. }
    \label{fig:bleu}
\end{figure}

% \begin{table}[ht]
% \centering
% \small
% \resizebox{\linewidth}{!}{
%   \begin{tabular}{l|cccccc}
%      \toprule
%      \multirow{2}{*}{} & \multicolumn{2}{c}{\textbf{AR}} & \multicolumn{2}{c}{\textbf{ZH}} & \multicolumn{2}{c}{\textbf{HI}} \\
%      & F1.$\uparrow$ & MET.$\downarrow$ & F1.$\uparrow$ & MET.$\downarrow$ & F1.$\uparrow$ & MET.$\downarrow$ \\
%      \midrule
%      TNT-14B    & {66.53} & - & {60.12} & - & 76.26 & - \\
%      TNT-7B     & 65.30 & 42.76 & 59.73 & 48.38 & 73.89 & 51.75 \\
%      TNT-3B     & 54.93 & 38.20 & 55.62 & 40.65 & 67.87 & 43.95 \\
%      TNT-1.5B   & 46.51 & 29.06 & 48.46 & 32.16 & 60.59 & 31.02 \\
%      \bottomrule
%   \end{tabular}
% }
%   \caption{Performance of our approach using models with different parameters.}
%   \label{tab:ablation}
% \end{table}

\begin{figure}[t]
    \centering
    \includegraphics[width=0.92\linewidth]{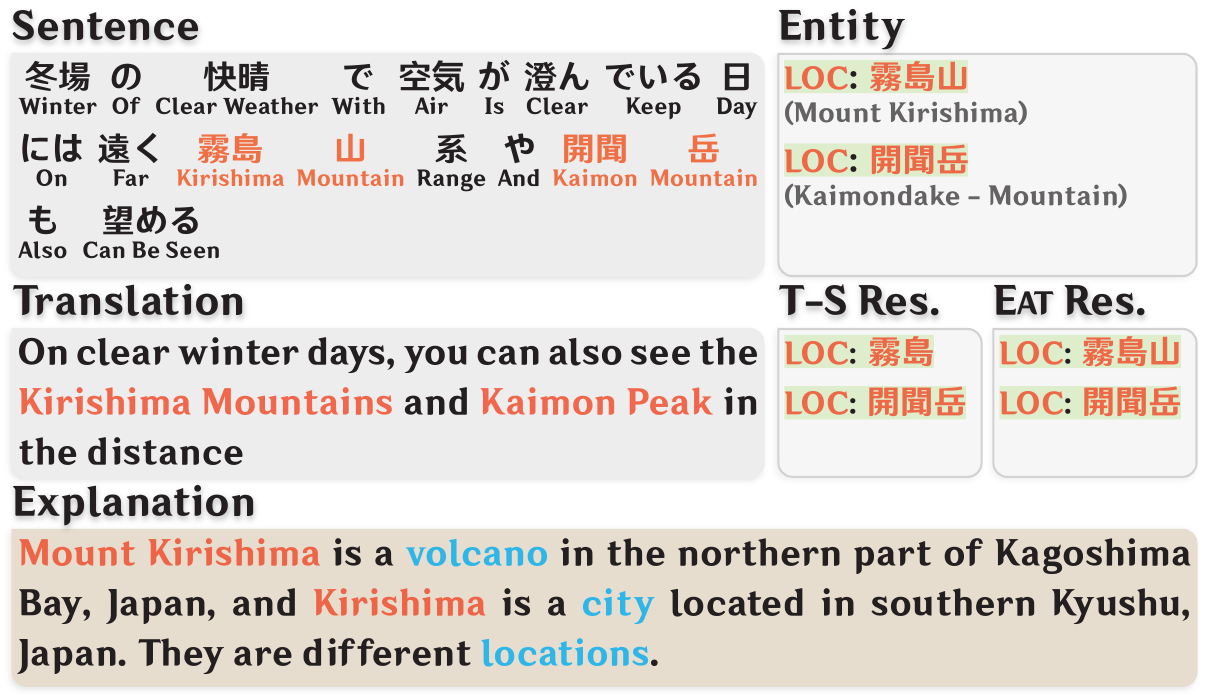}
    \caption{Comparison of the NER results for T-S based approaches (T-S Res.) and our approach (\textsc{Eat} Res.).}
    \label{fig:case1ja1}
\end{figure}

\begin{table}[t]
\centering
\small
\resizebox{0.88\linewidth}{!}{
  \begin{tabular}{l|cccc}
     \toprule
     {} & {\textbf{AR}} & {\textbf{HI}} & {\textbf{JA}} & {\textbf{ZH}} \\
     \midrule
     \textsc{Eat} w/o FT    & \textbf{66.53} & \textbf{76.26} & \textbf{45.43} & \textbf{60.12} \\
     GPT-4  & 60.60 & 56.14 & 34.44 & 52.35 \\
     Qwen-14B-ICL & 45.71 & 61.18 & 41.95 & 52.89 \\
     Qwen-14B-DA & 29.82 & 72.27 & 41.34 & 49.05 \\
     \bottomrule
  \end{tabular}
}
  \caption{Performance comparison of our approach and CoT-based LLMs for ZCL-NER. ICL denotes in-context learning for Qwen with source labeled data. DA denotes data augmentation for NER model with translated source labeled data. Details are described in Appendix~\ref{sec:icl}and~\ref{sec:da}.}
  \label{tab:ablation-gpt}
\end{table}

\begin{figure}[t]
    \centering
    \includegraphics[width=0.92\linewidth]{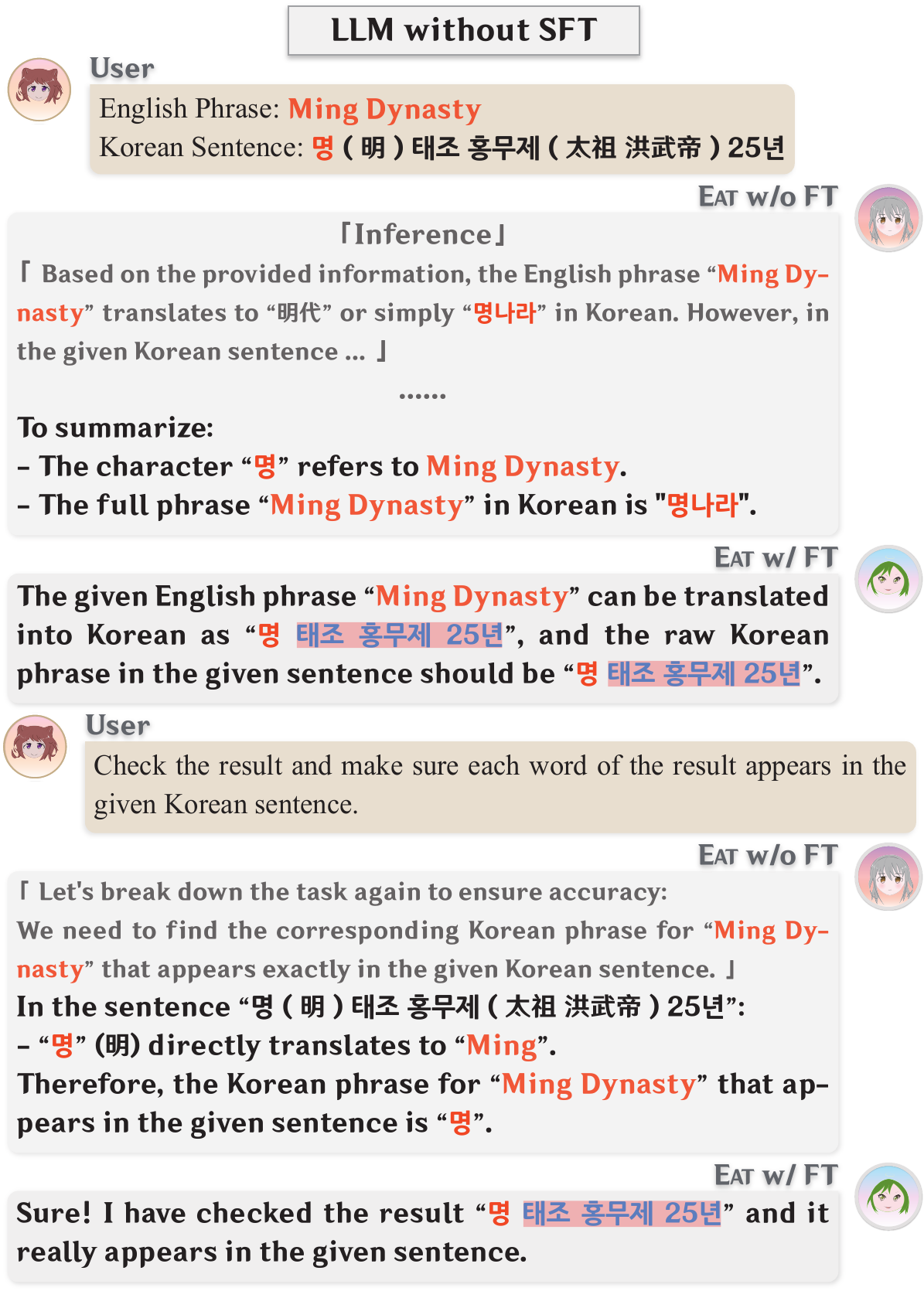}
    \caption{Comparison of the entity alignment process for LLM with and without FT. Texts in pink background denote the wrong alignments, w/o pink denotes right.}
    \label{fig:case2ko1}
\end{figure}

% \noindent $\bullet$ \textbf{Ablation Study}

\textbf{Impact of Translation Ability.} We present Figure \ref{fig:bleu} and Table \ref{tab:ablation} to examine \textit{whether translation ability directly affects NER performance by the translation quality evaluation metrics we introduced}. 

Figure~\ref{fig:bleu} shows that on each language, the coordinate point of the F1-Score is always between BLEU and Loss $L_e$. This suggests a positive correlation between NER performance and translation ability. Table~\ref{tab:ablation} demonstrates that the BLEU score decreases as the size of homologous LLMs decreases, which means that the model’s translation ability weakens. This, in turn, negatively impacts the performance of our \textsc{Eat} framework. These findings underscore the critical importance of improving translation ability to enhance ZCL-NER performance.
% This reinforces the importance of enhancing the translation ability for better ZCL-NER. 
In addition, we also evaluate our approach with heterologous LLMs (different backbones) as presented in Appendix~\ref{sec:more-res}.
% We evaluate the impact of LLMs' translation ability on cross-lingual NER. The model's multilingual ability differs due to different amounts of training resources for each language. As shown in Figure~\ref{fig:bleu}, the NER results generally follow the same trend as the translation ability. Note that the BLEU scores and entropy loss are co-relevant to the translation ability, and the NER results are also affected by the NER model.

% In addition, translation ability deteriorates as model parameters decrease~\cite{qwen2.5}. We adopt four Qwen2.5-Instruct~\cite{qwen2, qwen2.5} models with decreasing parameters as the backbone. As shown in Table~\ref{tab:ablation}, deterioration of translation ability reduces NER performance. 

\textbf{Impact of NER Extractor Ability.}
To better compare our approach with T-S based approaches, we leverage \textbf{mBERT}~\cite{mbert} with sequence-labeling as the NER model. As shown in Table~\ref{tab:ablation-ner}, there is a little difference between mBERT and T5 as both models are well-trained for the English NER task. In general, our \textsc{Eat} with T5 performs slightly better than it with mBERT. This suggests the robustness of our \textsc{Eat}.

\textbf{Impact of Direct Using LLMs for ZCL-NER.}
We leverage \textbf{both LSL and NSL LLMs} (GPT and Qwen) directly on the ZCL-NER task to explore whether LLMs could perform better without our proposed Dual-Translation mechanism. As shown in Table~\ref{tab:ablation-gpt}, our approach performs better than GPT-4~\cite{openai2024gpt4}, which is recognized as one of the best LLMs. Even directly using source-labeled data for Qwen in-context learning (\textbf{ICL}) or translated into target language (\textbf{DA}) both fail to improve its poor NER performance, which aligns with insights from existing studies~\cite{xie2023emnlp, xie2024acl, li2025m2ivefficientfinegrainedmultimodal, li2025taco}.

\begin{table}[t]
\centering
\small
\resizebox{0.96\linewidth}{!}{
  \begin{tabular}{l|cccc}
     \toprule
     {} & {\textbf{HI}} & {\textbf{KO}} & {\textbf{RU}} & {\textbf{ZH}} \\
     \midrule
     \textsc{Eat} w/o FT (Ours) & \textbf{47.38} & \textbf{64.12} & \textbf{52.66} & \textbf{63.00} \\
     GPT-4 & 47.02 & 49.45 & 37.72 & 48.58 \\
     DenKD~\cite{DenKD} & 33.67 & 44.61 & 45.26 & 41.48 \\
     \bottomrule
  \end{tabular}
}
  \caption{Performance comparison on MultiCoNER-1~\cite{malmasi-etal-2022-multiconer}.}
  \label{tab:mcn1}
\end{table}

\textbf{Generalization Ability.} As shown in Table~\ref{tab:mcn1}, our approach also outperforms the SOTA teacher-student model and GPT-4 significantly on MultiCoNER-1~\cite{malmasi-etal-2022-multiconer}. This suggests the excellent generalization ability of our proposed \textbf{\textsc{Eat}}.

\textbf{Case Study.}
% \label{sec:case}
From \textbf{NER Results} aspect, as shown in Figure~\ref{fig:case1ja1}, our approach \textsc{Eat} correctly comprehends the semantics of the given sentence due to accurate translation and explanation. Hence, \textsc{Eat} accurately identifies the entities and their corresponding tags. However, T-S based approaches struggle to grasp the meaning of all phrases. As a result, they recognize a wrong entity that mismatches the intended semantics.

From \textbf{Entity Alignment} aspect, as shown in Figure~\ref{fig:case2ko1}, \textsc{Eat} model w/ FT seems to become inflexible. It loses the ability of inference and thus outputs the wrong entity alignment. However, \textsc{Eat} model w/o FT can infer multiple alignment results for entities from the source language to the target language. This not only presents error analysis of our \textsc{Eat} w/ FT, but also indicates that the inference progress will enhance alignment results and makes the NER results more accurate. This finding also matches recent studies~\cite{deepseekai2025deepseekr1incentivizingreasoningcapability, kimiteam2025kimik15scalingreinforcement}: reasoning progress can improve LLMs' performance.

% \begin{table}[ht]
% \centering
% \small
%   \begin{tabular}{l|ccc}
%      \toprule
%       & {\textbf{AR}} & {\textbf{ZH}} & {\textbf{HI}} \\
%      \midrule
%      TNT-14B & {66.53} & {60.12} & 76.26 \\
%      TNT-7B & 65.30 & 59.73 & 73.89 \\
%      TNT-3B & 54.93 & 55.62 & 67.87 \\
%      TNT-1.5B & 46.51 & 48.46 & 60.59 \\
%      \bottomrule
%   \end{tabular}
%   \caption{Performance of our approach using models with different parameters.}
%   \label{tab:ablation}
% \end{table}

\section{Conclusion}
We introduce a novel entity-aligned translation (\textsc{Eat}) approach with LLMs for zero-shot cross-lingual NER (ZCL-NER) approach to mitigate the linguistic differences between non-Latin script language (NSL) and English, so as to better leverage the rich English NER resources for multilingual NER tasks.
% With our proposed dual translation mechanism, our approach achieves better performance on NSL than previous SOTA for ZCL-NER. 
In addition, we fine-tune the LLM using collected source-oriented cross-lingual corpora to enhance entity alignments for better NER. Furthermore, we employ BLEU and information entropy to analyze the correlation between NER performance and translation ability.

\section*{Acknowledgements}
This work was supported by NSFC grants (No. 62206193 and No. 62376178) and General Research Fund (GRF) project sponsored by the Research Grants Council Hong Kong (Project No.15611021).

\section*{Limitations}
Although our approach has achieved impressive results on zero-shot cross-lingual NER, there are still limitations. The LLMs are pre-trained on uneven corpora in different languages, making their multilingual ability differs. However, multilingual semantic understanding is crucial in our approach, as it directly affect translation ability. If the LLM fails to operate dual translation, our approach will also fail. In addition, the amount of knowledge in different languages correlates positively with the size of LLMs, and the model size affects computational resources usage. For instance, Qwen2.5 supports 29 languages while GPT-4 supports over 80,
% and their requirements for resources are quite different.
with significantly different resource requirements.
Therefore, we must strike a balance between performance and universality, where the LLM is large enough to undertake the multilingual dual translation while remaining as small as possible to maximize the inference speed and minimize the energy usage and carbon emissions.

% \section*{Acknowledgements}
% PENDING

\bibliography{custom}

\appendix

\section*{Appendix}

\section{Evidences of our Motivation}
\begin{figure}[t]
    \centering
    \includegraphics[width=\linewidth]{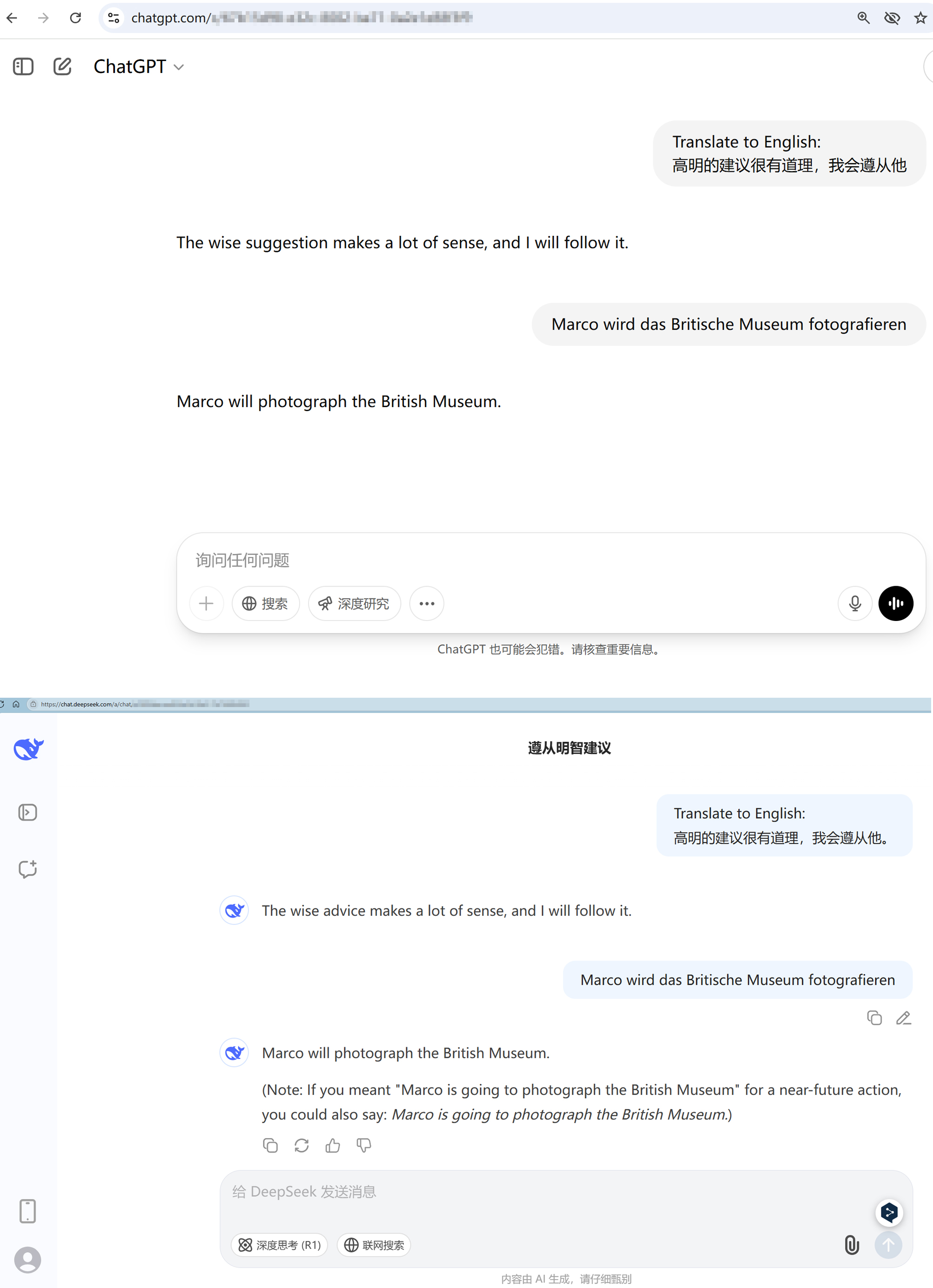}
    \caption{The screenshots of the examples in Figure~\ref{fig:lexicon}. The left is ChatGPT-4o and the right is Deepseek-V3.}
    \label{fig:evidence-intro}
\end{figure}

As the case in Figure~\ref{fig:lexicon}, the Person entity ``高明'' in target language (Chinese) is incorrectly translated as the adjective ``clever'' by GPT-4o and Deepseek \textbf{tested in May 2025}. This can be seen from Figure \ref{fig:evidence-intro} as evidences. Such translation inconsistencies make it impossible to proceed our task, naturally motivating this work.

\section{Details of Method}
\subsection{Task Formulation}
\begin{figure*}[t]
    \centering
    \includegraphics[width=\linewidth]{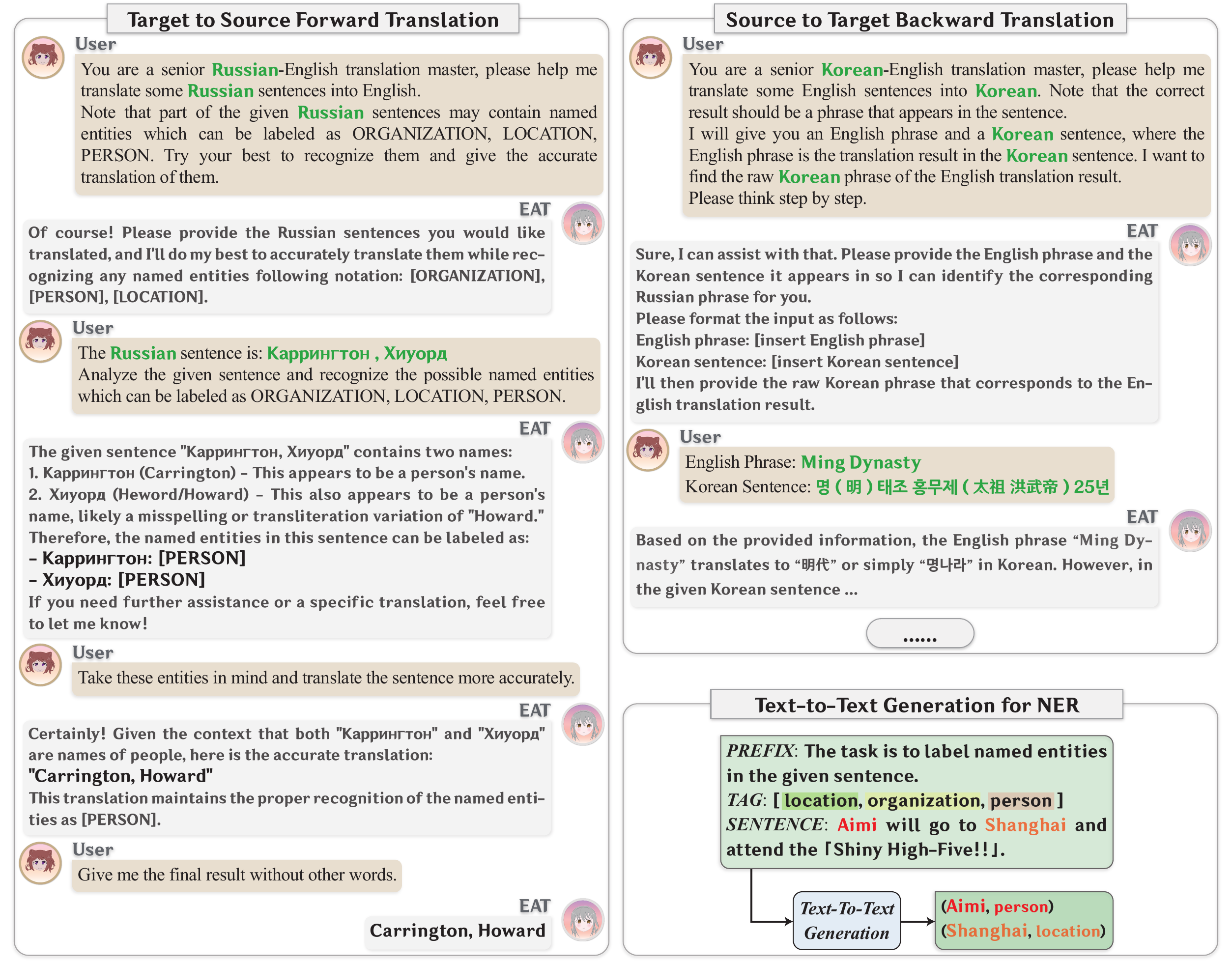}
    \caption{Full process of Dual Translation for ZCL-NER. Texts in \textcolor[RGB]{66,185,131}{green} denote the interchangeable templates. The portion which has been appeared in Figure~\ref{fig:case2ko1} is omitted in Source to Target Backward Translation.~\protect\footnotemark[1]}
    \label{fig:case-full}
\end{figure*}

Given an $n$-token sentence $\boldsymbol{x} = <x_1,\cdots,x_n>$ and $k$-type entity set $\mathbb{T} = <t_1,\cdots,t_k>$, the object of NER task is to extract all entities $\boldsymbol{e}_i \in \boldsymbol{E}$ from $\boldsymbol{x}$ and assign one of the types in $\mathbb{T}$ to each entity, where $\boldsymbol{e}_i = (\boldsymbol{x}_{start:end}, t)$ denotes the $i$-th entity of $\boldsymbol{x}$ and $t \in \mathbb{T}$ refers to the type of the entity. $\boldsymbol{x}_{start:end}$ refers to a continues word span $<x_{start},\cdots,x_{end}>$ in $\boldsymbol{x}$, where $start$ and $end$ refers to the entity boundary indexes respectively. Given dataset $\mathcal{D}_s$ of the source language (i.e., English in our setting) and dataset $\mathcal{D}_t$ of the target language, the objective of the ZCL-NER task is to acquire target-related knowledge from $\mathcal{D}_s$ to enhance model's performance on $\mathcal{D}_t$.

\subsection{Detailed Process of Dual Translation for ZCL-NER}
To better describe our proposed \textsc{Eat} framework, we give an example in Figure~\ref{fig:case-full} of the key process: Target to Source Forward Translation, Text-to-Text Generation for NER, and Source to Target Backward Translation.

\noindent \textbf{Prompts} Since our approach is not focusing on prompt engineering, we just use the only prompts shown in Figure~\ref{fig:case2ko1} and \ref{fig:case-full}.

\footnotetext[1]{Part of the images are provided by Gu:~\url{https://b23.tv/RAKSxQg}}

\subsection{Details of Entity-level Alignment Fine-tuning}
\label{sec:sft-app}
To amplify the DT model's entity-level alignment ability, we leverage \textsc{Eacl} Corpora obtained above to fine-tune the model. Specifically, we leverage Quantized-LoRA(QLoRA)~\cite{QLORA} to accelerate model fine-tuning and reduce memory usage under constrained resources.
Instead of global quantization, we use block-wise $k$-bit quantization~\cite{iclr/DettmersLSZ22} for its improvements in terms of accuracy, efficiency, and flexibility. More formally, for a given tensor $\boldsymbol{T}$, it is chunked into ${n}$ contiguous blocks and each block is flattened to $[-1,1]$:
% \begin{equation}
%     Chunk(\boldsymbol{T}) = [\boldsymbol{T}_1,\cdots,\boldsymbol{T}_n], \boldsymbol{T}_i \xrightarrow{Norm.} [-1, 1]
% \end{equation}
\begin{equation}
    Q_b(\boldsymbol{T}) = [\boldsymbol{T}_1^q,\cdots,\boldsymbol{T}_n^q]
\end{equation}
The $i$-th block is independently quantized as:
\begin{equation}
    \begin{split}
        {\boldsymbol{T}_i^q} = Q(\boldsymbol{T}_i) & = round(\frac{2^{k-1}-1}{absmax(\boldsymbol{T}_i)}\boldsymbol{T}_i) \\
        & = round({c_i} \cdot \boldsymbol{T}_i)
    \end{split}
\end{equation}
where $c_i \in \boldsymbol{c}$ is the quantization constant for each block. And the dequantization is:
\begin{equation}
    \boldsymbol{T}_i = D({c}_i, \boldsymbol{T}_i^q) = \frac{\boldsymbol{T}_i^q}{c_i}
\end{equation}

The model is quantified in $4$-bit NormalFloat (NF4) as described above, and the quantization constants $\boldsymbol{c}$ are quantized as $\boldsymbol{c}^d$ to further reduce memory usage. For a single layer's parameters of the model with LoRA adapter is: 
\begin{equation}
    \boldsymbol{Y} = \boldsymbol{X}{D}^2 (\boldsymbol{c}, \boldsymbol{c}^d, \boldsymbol{W}^q) + \boldsymbol{X}\boldsymbol{L}_1\boldsymbol{L}_2
\end{equation}
\begin{equation}
    {D}^2(\boldsymbol{c}, \boldsymbol{c}^d, \boldsymbol{W}^q) = D(D(\boldsymbol{c}, \boldsymbol{c}^d), \boldsymbol{W}^q) = \boldsymbol{W}
\end{equation}
where $\boldsymbol{W}$ and $\boldsymbol{W^q}$ denote the model's raw and quantized weights, and $\boldsymbol{L}_1\boldsymbol{L}_2$ denotes the trainable parameters of the LoRA adapters.

Cross-entropy loss is optimized to train the DT model:
\begin{equation}
    \boldsymbol{y} = \textsc{Lm}(e, {v}_1^a)
\end{equation}
\begin{equation}
    L_T(\boldsymbol{y},\boldsymbol{\hat{y}}) = -\sum_{i=1}^{\eta}\hat{y}_i\log (y_i)
\end{equation}
where $\hat{y}_i \in \boldsymbol{\hat{y}}$ is $u^a$ in $(u^a,v_1^a)_e$ of \textsc{Eacl}. $y_i \in \boldsymbol{y}$ denotes the predicted entity. $\textsc{Lm}$ denotes a large language model and $\eta$ denotes the max length of model output.

\subsection{Details of Entity-aligned Translation Ability Evaluation}
% \label{sec:bleu}
We use BLEU score~\cite{bleu} and Information Entropy~\cite{ie} to measure the information loss in the translation process, so as to demonstrate the relevance between the NER results and the information loss.

\textbf{Bilingual Evaluation Understudy} (BLEU) is widely used as an evaluation metric in machine translation. BLEU is a fast and unified metric, and it can evaluate all languages effectively. Since there are no reference translations of source language $a$ for target language $b$, we obtain the candidate as:

Given a sentence $\boldsymbol{s}_r$ of target language $b$, the candidate sentence is:
\begin{equation}
    \boldsymbol{s}_c = \mathcal{T}^t_{a\rightarrow{b}}(\mathcal{T}^t_{b\rightarrow{a}}(\boldsymbol{s}_r))
\end{equation}
And the $n$-gram precision is:
\begin{equation}
    \mathcal{P}_n = \frac{\sum_{k}^{K_n}\min (h_k(\boldsymbol{s}_c), h_k(\boldsymbol{s}_r))}{\sum_{k}^{K_n} h_k(\boldsymbol{s}_c)}
\end{equation}
where $K_n$ denotes the $n$-gram divided sequence, and $h_k(\cdot)$ denotes the counts of $k$-th $n$-gram.

The \textbf{Brevity Penalty} is introduced to avoid the scoring bias:
\begin{align}
    \mathop{BP} = \left\{
    \begin{aligned}[c]
        1 \hspace{2.5em} &\mathrm{if} \hspace{1em} l_c > l_r \\
        e^{1-l_r / l_c} \hspace{1em} &\mathrm{if} \hspace{1em} l_c \leq l_r
    \end{aligned}
    \right.
\end{align}
where $l_c$ and $l_r$ are the lengths of $\boldsymbol{s}_c$ and $\boldsymbol{s}_r$.

We adopt $n=4$, and the BLEU score is finally calculated by the geometric mean of the $n$-gram precision:
\begin{equation}
    \mathop{BLEU} = \mathop{BP}\times \exp (\frac{1}{4}\sum_{n=1}^4 \log(\mathcal{P}_n))
\end{equation}

\textbf{Information Entropy} (Shannon Entropy) is also commonly used to quantify the information of the sentences. We leverage the Bi-Gram Model to calculate the joint information entropy as:
\begin{align}
    \begin{aligned}
    H(\boldsymbol{s}) &= \sum_{i=1}^{n}H(s_{i-1}, s_{i}) \\
                      &= \sum_{i=1}^{n} P(s_{i-1}, s_{i}) (-\log P(s_{i-1}|s_{i}))
    \end{aligned}
\end{align}
where $P(s_{i-1}, s_{i})$ denotes the joint probability of $s_{i-1}$, $s_{i}$ appearing in the $n$-length text $\boldsymbol{s}$ with $s_{i-1}$ exactly before $s_{i}$, and $P(s_{i-1}|s_{i})$ denotes the conditional probability of $s_{i-1}$ appearing before $s_{i}$.

We demonstrate the information loss is relevant to the entropy loss. The entropy loss $L_e$ is defined using the above Shannon Entropy $H(\cdot)$ as:
\begin{equation}
    L_e = \frac{H(\mathcal{T}_{b\rightarrow{a}}^t(\boldsymbol{s}_r))}{H(\boldsymbol{s}_r)}
\end{equation}
where $\boldsymbol{s}_r$ denotes the sentence of target language $b$ and $a$ denotes the source language.

\section{More Implementation Details}
As shown in Figure~\ref{fig:structure}, we first collect the \textsc{Eacl} Corpora as described in Section~\ref{sec:et-corpus}. The number of original English entities we used in the CoNLL2003 train set is 8,082. After obtaining all entity-text pairs, we remove those pairs where the entity does not appear in the corresponding text. Then the texts and entities are constructed in ShareGPT format following previous work~\cite{llamafactory}. Detailed statistics of our collected corpora are listed in Table~\ref{tab:etcor-detail}.

\subsection{Prompts}
All prompts used in our proposed \textbf{\textsc{Eat}} are listed in Figure~\ref{fig:case2ko1} and \ref{fig:case-full}. It is worth noting that these prompts are only normal descriptions and instructions of what we want LLMs to do, which do not require special design or strict screening. Therefore, we believe that there is no need to conduct experiments based on the same meaning but different forms of prompts.
\begin{table}[t]
\centering
\small
% \resizebox{\linewidth}{!}{
  \begin{tabular}{l|cccc}
     \toprule
     \multirow{2}*{\textbf{Lang.}} & \multicolumn{3}{c}{Tokens} & {\multirow{2}*{Ratio}} \\
      & Train & Valid & Test & \\
     \midrule
     DE & 195,387 & \ \ 97,805 & \ \ 97,646 & \ \ 1.75\% \\
     ES & 129,283 & \ \ 64,329 & \ \ 64,728 & \ \ 7.30\% \\
     FR & 136,788 & \ \ 68,220 & \ \ 68,754 & \ \ 4.07\% \\
     NL & 169,449 & \ \ 84,146 & \ \ 85,122 & \ \ 0.39\% \\
     \midrule
     EN & 160,394 & \ \ 80,536 & \ \ 80,326 & 19.76\% \\
     \bottomrule
  \end{tabular}
  % }
  \caption{Detailed information of our selected languages in the dataset. Ratio represents the approximate proportion of speakers to the total world population, and the statistics are referenced from Wikipedia. }
  \label{tab:cor-latin-detail}
\end{table}

\begin{table*}[t]
  \centering
  \small
  % \resizebox{\linewidth}{!}{
  \begin{tabular}{l|cccc}
     \toprule
     \multirow{2}*{} & \multicolumn{4}{c}{\textbf{Latin Scripts}} \\
          & \textbf{DE} & \textbf{ES} & \textbf{FR} & \textbf{NL} \\
     \midrule
     mBert~\cite{FTDT}          & 78.64 & 74.55 & 80.20 & 82.55 \\
     CROP~\cite{crop}  & 60.87 & 62.05 & 55.79 & 54.77 \\
     \midrule
     TSLM~\cite{Single-TS}      & 79.96 & 77.18 & 80.38 & 83.80 \\
     RIKD~\cite{RIKD}           & 80.20 & 77.79 & 81.20 & 84.65 \\
     AdvPicker~\cite{AdvPicker} & 79.72 & 77.81 & 79.91 & 84.27 \\
     % MulDA~\cite{mulda}         & 53.62 & 67.46 &       & 37.05 & 67.68 & 52.69 & 65.00 & 41.77 &       \\
     DualNER~\cite{DualNER}     & 80.17 & 78.42 & 80.92 & 84.36 \\
     MSD~\cite{MSD}             & 80.62 & 75.75 & 81.16 & 84.23 \\
     ProKD~\cite{ProKD}         & 79.74 & 79.19 & 81.45 & \underline{84.73} \\
     DenKD~\cite{DenKD}         & \underline{82.50} & \textbf{84.68} & 82.34 & \textbf{85.69} \\
     \midrule
     \textsc{Eat}                       & \textbf{83.91} & 83.43 & \underline{84.50} & 81.79 \\
     \textsc{Eat} w/ SFT                & 81.69 & \underline{84.35} & \textbf{85.22} & 81.69 \\
     \bottomrule
  \end{tabular}
  % }
  \caption{Performance comparison of our approaches on languages using Latin scripts. Bold represents the best result, and underlining represents the second best result.}
  \label{tab:addi-wiki}
\end{table*}

\subsection{Baselines}
To compare our approach with previous translation-based and T-S framework based approaches, we report the baselines as follows:

\textbf{Translation based:}

1) \textbf{mBert}~\cite{FTDT} leverages a pre-trained model to directly transfer from source languages to target languages.
% 2) \textbf{CROP}~\cite{crop} leverages a sequence translation model with a back-translation schema for cross-lingual NER. We modify this approach as m-\textbf{CROP} due to the defects in the back-translation schema, and detailed information can be found in Appendix~\ref{sec:mo-crop}.

2) \textbf{Awesome-align}~\cite{Awesome-align} fine-tunes PLMs with paralleled data on source and target languages to extract label alignments.

3) \textbf{CROP}~\cite{crop} leverages a sequence translation model to operate the ZCL-NER task with a cross-lingual entity projection framework.

4) \textbf{EasyProject}~\cite{EasyProject} improves mark-then-translate method to better perform translation and label projection.

5) \textbf{CLaP}~\cite{clap} proposes contextual translation to better translate the labels to the target languages.

\textbf{Teacher-Student Framework based:}

3) \textbf{TSLM}~\cite{Single-TS} proposes vanilla teacher-student learning to distill knowledge for cross-lingual NER.

4) \textbf{RIKD}~\cite{RIKD} proposes a teacher-student learning approach with reinforcement-learning-based knowledge distillation.

5) \textbf{AdvPicker}~\cite{AdvPicker} introduces adversarial learning in the training process of the teacher model to denoise in knowledge distillation.

6) \textbf{DualNER}~\cite{DualNER} proposes a unified framework that combines NER learning paradigms and applies multi-task learning for knowledge distillation.

7) \textbf{MSD}~\cite{MSD} designs a multichannel distillation framework with a parallel domain adaptation to efficiently transfer information.

8) \textbf{ProKD}~\cite{ProKD} proposes prototypical alignment with prototypical self-training for knowledge distillation to better acquire knowledge.

9) \textbf{DenKD}~\cite{DenKD} proposes a denoising approach using uncertainty- and discrepancy-awareness to reduce the noise in the knowledge distillation process, which is the \emph{SOTA model}.

\begin{table}[t]
\centering
\small
  \resizebox{\linewidth}{!}{
  \begin{tabular}{l|cccccc}
     \toprule
     \multirow{2}{*}{} & \multicolumn{2}{c}{\textbf{DE}} & \multicolumn{2}{c}{\textbf{ES}} & \multicolumn{2}{c}{\textbf{NL}} \\
     & Qwen & Llama & Qwen & Llama & Qwen & Llama \\
     \midrule
     \textsc{Eat}-14B & 77.51 & - & 82.84 & - & 78.76 & - \\
     \textsc{Eat}-7B$^*$ & 71.39 & 60.16 & 78.36 & 62.06 & 76.24 & - \\
     \textsc{Eat}-3B & 56.05 & 48.20 & 56.90 & 49.33 & 58.46 & - \\
     \textsc{Eat}-1B$^*$ & 40.31 & 19.16 & 40.60 & 15.71 & 40.39 & - \\
     \bottomrule
  \end{tabular}
  }
  \caption{Performance comparison on CoNLL dataset using different models as backbones. '-' represents no result since Dutch is not officially supported by Llama3 and Llama3 does not have an official 14B version.}
  \label{tab:addi-conll}
\end{table}

\begin{table}[t]
\centering
\small
\resizebox{\linewidth}{!}{
  \begin{tabular}{l|cccc}
     \toprule
     {} & {\textbf{HI}} & {\textbf{KO}} & {\textbf{RU}} & {\textbf{ZH}} \\
     \midrule
     \textsc{Eat} w/o FT (Ours) & \textbf{47.38} & \textbf{64.12} & \textbf{52.66} & \textbf{63.00} \\
     GPT-4 & 47.02 & 49.45 & 37.72 & 48.58 \\
     DenKD~\cite{DenKD} & 33.67 & 44.61 & 45.26 & 41.48 \\
     \bottomrule
  \end{tabular}
}
  \caption{Performance comparison of our approach with previous SOTA approaches on MultiCoNER-1. Due to the limit of computational resources and time, the test set is downsized through random sampling.}
  \label{tab:mcn}
\end{table}

\begin{table}[t]
\centering
\small
% \resizebox{\linewidth}{!}{
  \begin{tabular}{l|cc}
     \toprule
     {} & {\textbf{ZH}} & {\textbf{AR}} \\
     \midrule
     5 rounds & {61.33} & {66.58} \\
     3 rounds & 61.27 & 67.29 \\
     1 round & 59.62 & 64.17 \\
     \bottomrule
  \end{tabular}
% }
  \caption{Performance comparison of our approach with different CoT rounds.}
  \label{tab:ablation-round}
\end{table}

\begin{figure}[t]
    \centering
    \includegraphics[width=0.9\linewidth]{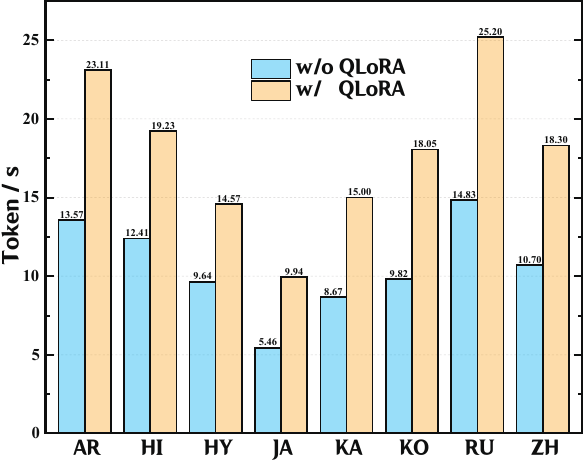}
    \caption{Speed of token generation on different languages.}
    \label{fig:efficiency}
\end{figure}

\section{Additional Results}
\label{sec:more-res}
\subsection{Results on Latin Script Languages}

We also conduct experiments on LSL for better comparison, including German (DE), Spanish (ES), French (FR), and Dutch (NL). As shown in Table~\ref{tab:addi-wiki}, our approach achieves similar performance compared with previous teacher-student learning models. This illustrates our approach works on not only NSL but also LSL.

\subsection{Results on CoNLL dataset}
To better evaluate our approach, we conduct experiments on the CoNLL dataset and use \textbf{Llama3-Instruct}~\cite{llama3} to compare. As shown in Table~\ref{tab:addi-conll}, Qwen2.5 defeats Llama3 with the same level parameters on all languages. This result matches the translation ability difference evaluated in \citet{qwen2.5}.

\subsection{Results on MultiCoNER-1}
To better evaluate the generalization ability of our approch, we conduct experiments on MultiCoNER-1~\cite{malmasi-etal-2022-multiconer}. MultiCoNER-1 is a larger dataset and all the annotations are made by human. We mainly select four most representative NSLs and two SOTA approaches for fair comparison, as shown in Table~\ref{tab:mcn}. From this table, we can see that our approach obviously outperforms the SOTA teacher-student model and GPT-4. This reveals the excellent generalization ability of our proposed approach \textbf{\textsc{Eat}}.

\subsection{Ablation Study on CoT Rounds}
To measure round differences, we add ablation studies about the round depth. As shown in Table~\ref{tab:ablation-round}, the results of round 1 were not good, while the results of round 5 were too time-consuming and unstable. Therefore, 3 rounds should be a realistic choice..

\subsection{Inference Acceleration of QLoRA}
We evaluate the efficiency of QLoRA described in Section~\ref{sec:sft}. As shown in Figure~\ref{fig:efficiency}, the token generation speed increases after using QLoRA.

% \begin{table}[t]
% \centering
% \small
%   \begin{tabular}{l|c}
%      \toprule
%      {} & {\textbf{HI}} \\
%      \midrule
%      \textsc{Eat} w/o FT    & {76.26} \\
%      DA  & 72.27 \\
%      \bottomrule
%   \end{tabular}
%   \caption{Performance comparison of our approach and DA approach for ZCL-NER.}
%   \label{tab:ablation-da}
% \end{table}
\section{Comparison with In Context Learning}
\label{sec:icl}
We conduct experiments on in context learning (ICL) for ZCL-NER. We use English NER examples and ask the model to do NER task following the given examples.

\section{Comparison with Data Augmentation}
\label{sec:da}
We conduct experiments on data augmentation (DA) for ZCL-NER. We translate the English (EN) train set into the target languages, and use translated train set to train the model for NER task.

The EN train set is directly translated using Qwen2.5-14B-Instruct without fine-tuning. Since T5-base can not directly take non-Latin scripts such as Devanagari and Hanzi as input, we use \textbf{mT5-base}~\cite{mt5} as an alternative. The model is trained for 50 epoch with a learning rate of ${1.0e{-4}}$.

The NER results are evaluated as described in Section~\ref{sec:expe}. As shown in Table~\ref{tab:ablation-gpt}, the DA approach performs worse than our approach. Furthermore, it also performs worse than the best T-S based approach.

In addition to the poor performance, no matter how many sentences are recognized, it is necessary for the DA approach to translate the full EN train set into one specific language for the NER model training. In other words, even recognizing just one sentence also requires complete translation and training. Translation of the full EN train set and training the NER model require large amounts of computational resources, as well as time consumption. However, the minimum train requirement of our approach is just the English NER model. Our approach is more flexible and efficient than DA approach, and is more practical for real-world applications.

\section{Fine-tuning Trade-offs}
In our preliminary experiments, the loss reaches a stable convergence state after about 5 epochs of fine-tuning. Therefore, we set fine-tuning epochs to 5. To this end, we add further ablation studies on performance with different EACL corpus sizes. 

\begin{table}[t]
\centering
\small
% \resizebox{\linewidth}{!}{
  \begin{tabular}{l|cc}
     \toprule
     {} & {\textbf{ZH}} & {\textbf{AR}} \\
     \midrule
     Full size & {61.27} & {67.29} \\
     1k samples & 61.20 & 67.10 \\
     \bottomrule
  \end{tabular}
% }
  \caption{Performance comparison of our approach with different EACL size. 1k samples are randomly selected from the full corpus.}
  \label{tab:tradeoff}
\end{table}

As shown in Table~\ref{tab:tradeoff}, smaller size of EACL corpus leads to the slight drop on performance. This suggests that the data scale and fune-tuning epochs we are currently using are reasonable.

\begin{table}[t]
\centering
\small
% \resizebox{\linewidth}{!}{
  \begin{tabular}{l|cc}
     \toprule
     {} & {\textsc{Eat}} & {DenKD} \\
     \midrule
     Avg. second per iterator & {5.85} & {1.86} \\
     Avg. second per token & 0.026 & 0.74 \\
     \bottomrule
  \end{tabular}
% }
  \caption{Average time comsumption per iterator or token for \textsc{Eat} and DenKD~\cite{DenKD}.}
  \label{tab:overhead}
\end{table}

\section{Inference Cost and Computational Overhead Comparison}
We add the average time consumption of a single sentence or token inference for assessment of practical feasibility. 

Table~\ref{tab:overhead} indicates that our EAT consumes slightly more time than traditional best-performed method. But this is due to the time-consuming inference process of using LLMs, and our EAT performs much better than DenKD.
Following the description in the paper of DenKD, the algorithm complexity of DenKD is $O(n\log n)$ (ignoring MLPs). During training strategy, for each input token, DenKD needs to calculate the Prediction Discrepancy Loss with a double circulation. However, the algorithm complexity of EAT is $O(n)$, as there is no additional calculations for loss.

\end{CJK}
\end{document}